\documentclass[runningheads]{llncs}
\usepackage[T1]{fontenc}
\usepackage{graphicx}
\usepackage{booktabs}
\usepackage[misc]{ifsym}
\newcommand{\corr}{(\Letter)}
\usepackage{microtype}
\usepackage{graphicx}
\usepackage{enumerate}
\usepackage{subcaption}
\usepackage{algorithmic}
\usepackage[ruled,vlined]{algorithm2e}

\usepackage{booktabs} %

\usepackage{hyperref}

\usepackage{mathtools}

\usepackage{amsmath,amssymb,amsfonts}
\usepackage{amsfonts} 
\usepackage{graphicx}
\usepackage{booktabs} %

\usepackage[numbers,sort&compress]{natbib}
\usepackage{wrapfig}

\usepackage{ulem}

\usepackage{tabularx}
\usepackage{lipsum}
\usepackage{newfloat}
\usepackage{listings}
\usepackage{stfloats}
\usepackage[capitalize,noabbrev]{cleveref}

\usepackage[table,xcdraw,dvipsnames]{xcolor}
\definecolor{mine}{RGB}{205, 232, 248}

\hypersetup{colorlinks,linkcolor={red},citecolor={green}}

\newtheorem{assumption}{Assumption}
\crefname{assumption}{assumption}{assumptions}

\usepackage[textsize=tiny]{todonotes}
\usepackage{thmtools, thm-restate}

\usepackage[misc]{ifsym}
\usepackage{lipsum}
\begin{document}

\title{Pareto Multi-Objective Alignment \\ for Language Models}

\titlerunning{Pareto Multi-Objective Alignment for Language Models}
\author{Qiang He \corr \and
Setareh Maghsudi } 

\institute{Ruhr University Bochum, 44801 Bochum, Germany \\
\email{\{qiang.he, setareh.maghsudi\}@rub-uni-bochum.de}}
\tocauthor{Qiang He, Setareh Maghsudi}
\toctitle{Pareto Multi-Objective Alignment for Language Models}

\authorrunning{Q. He and S. Maghsudi}
\maketitle

\begin{abstract}
Large language models (LLMs) are increasingly deployed in real-world applications that require careful balancing of multiple, often conflicting, objectives, such as informativeness versus conciseness, or helpfulness versus creativity. However, current alignment methods, primarily based on reinforcement learning from human feedback (RLHF), optimize LLMs toward a single reward function, resulting in rigid behavior that fails to capture the complexity and diversity of human preferences. This limitation hinders the adaptability of LLMs to practical scenarios, making multi-objective alignment (MOA) a critical yet underexplored area.
To bridge this gap, we propose \underline{PA}reto \underline{M}ulti-Objective \underline{A}lignment (PAMA), a principled and computationally efficient algorithm designed explicitly for MOA in LLMs. In contrast to computationally prohibitive gradient-based multi-objective optimization (MOO) methods, PAMA transforms multi-objective RLHF into a convex optimization problem with a closed-form solution, significantly enhancing scalability. Traditional gradient-based MOO approaches suffer from prohibitive $\mathcal{O}(n^2d)$ complexity, where $d$ represents the number of model parameters, typically in the billions for LLMs, rendering direct optimization infeasible. PAMA reduces this complexity to $\mathcal{O}(n)$ where $n$ is the number of objectives, enabling optimization to be completed within milliseconds.
We provide theoretical guarantees that PAMA converges to a Pareto stationary point, where no objective can be improved without degrading at least one other. Extensive experiments across language models ranging from 125M to 7B parameters demonstrate PAMA's robust and effective multi-objective alignment capabilities, consistently outperforming baseline methods, aligning with its theoretical advantages. PAMA provides a highly efficient solution to the MOA problem that was previously considered intractable, offering a practical and theoretically grounded approach to aligning LLMs with diverse human values, paving the way for versatile and adaptable real-world AI deployments.
\end{abstract}

\section{Introduction}
Large language models (LLMs) have demonstrated impressive capabilities across diverse natural language tasks~\citep{transformer,gpt1,bert}, receiving significant attention from academia and industry~\citep{llama2,gpt4}. However, a critical deployment challenge is aligning LLMs with complex human values. Currently, reinforcement learning from human feedback (RLHF) is the predominant alignment approach~\citep{rlhf,hh-rlhf}, fine-tuning models against a single reward function that approximates human preferences practically~\citep{llama2,deepseek,llama3}. While effective in producing coherent outputs, this single-objective alignment severely restricts LLMs, resulting in homogeneous behaviors that fail to reflect the diverse spectrum of human values.

Real-world scenarios increasingly demand models that simultaneously balance multiple, often conflicting objectives, such as informativeness versus conciseness, helpfulness versus creativity, and etc~\citep{he2025moc,llama2,deepseek}. Therefore, aligning LLMs requires moving beyond single-objective reward models towards multi-objective alignment (MOA), which considers multiple and potentially conflicting reward signals~\citep{ric,rewardedsoup}. Despite recent interest, a theoretically grounded and practical method for achieving MOA in LLMs has yet to be established.

A naive solution aggregates heterogeneous rewards into a single scalar objective~\citep{morl-survey}, but this simplification neglects inherent reward conflicts, often leading to biased or misaligned outcomes~\citep{boyd2004convex}. Existing gradient-based multi-objective optimization (MOO) methods~\citep{MGDA,pcgrad,cagrad,ica} are also impractical for large-scale LLMs due to prohibitively expensive gradient computations. For instance, MGDA~\citep{MGDA} involves min-norm operations with time complexity~$\mathcal{O}(n^2d)$, making it infeasible for models with billions of parameters (e.g., $d=7$ billion). Thus, developing a scalable and principled MOA algorithm specifically for LLMs remains crucial.

In this work, we propose \underline{PA}reto \underline{M}ulti-Objective \underline{A}lignment (PAMA), a novel, computationally efficient algorithm designed explicitly for multi-objective alignment in LLMs. PAMA converts multi-objective RLHF into a convex optimization problem with a closed-form solution, eliminating expensive gradient calculations. Remarkably, PAMA achieves computational costs comparable to standard single-objective PPO algorithms, enabling efficient fine-tuning of 7-billion-parameter models on a single NVIDIA A6000 GPU. Unlike traditional methods~\citep{MGDA,cagrad} with $\mathcal{O}(n^2d)$ complexity, PAMA scales linearly with the number of objectives $\mathcal{O}(n)$, drastically reducing computational demands and enabling practical use with LLMs. For instance, when $n=10$ and $d=10^{10}$, existing approaches would require roughly $10^{12}$ computations, whereas PAMA completes the task in just 10 steps, demonstrating an exponential improvement in efficiency. In such an LLM setting, methods like MGDA~\citep{MGDA}, PCGrad~\citep{pcgrad}, and CAGrad~\citep{cagrad} become computationally infeasible, whereas PAMA remains tractable and scalable.

Furthermore, we provide theoretical guarantees of convergence to a Pareto stationary point, ensuring no single objective can improve without degrading others. To our knowledge, PAMA is the first theoretically grounded MOA algorithm for LLMs.

The theoretical advantages of PAMA are also reflected in our empirical results. Empirical evaluations validate PAMA across language models ranging from 125M to 7B parameters. Our experiments comprehensively demonstrate PAMA's robust and consistent superiority, while other baselines fail with large performance gaps. The results highlight PAMA's effectiveness, scalability, and robustness, aligned with its theoretical properties.

Our contributions are summarized as follows:
\begin{itemize}
    \item Pareto Multi-Objective Alignment: A novel and efficient multi-objective alignment algorithm for LLMs, reducing computational complexity from $\mathcal{O}(n^2d)$ to $\mathcal{O}(n)$, enabling efficient large-scale training.
    \item Theoretical Guarantees: We prove convergence of PAMA to a Pareto stationary point.
    \item Empirical Validation: Extensive experiments demonstrate PAMA's superior performance across multiple settings.
\end{itemize}

\section{Method~\label{sec: method}}
This section presents our approach to multi-objective alignment in the context of LLMs. We begin by formulating the problem and introducing Noon PPO, a variant of PPO~\citep{PPO}. We then propose PAMA, an algorithm designed to align LLMs with multiple objectives while ensuring convergence to a Pareto stationary point with theoretical guarantees.
\subsection{Problem Formulation}
RLHF consists of two main phases: reward modeling and policy optimization. In reward modeling, a reward function is trained on preference data to maximize the objective: $\mathcal{L}_{RM} = \mathbb{E}_{(x, y^w, y^l) \sim \mathcal{D}} [\log(\sigma(r(x, y^w) - r(x, y^l)))]$,
where, $ y^w $ and $ y^l $ denote the preferred and less desirable responses, respectively, $ x $ represents the prompt, and $ \sigma(\cdot) $ is the sigmoid function. 
In policy optimization, RLHF typically employs PPO to refine the policy by solving:
\begin{equation*}
    \arg\max_{\pi(y|x;\theta)} \mathbb{E}_{x \sim \mathcal{D}, y \sim \pi(\cdot|x)} \left[ r(x, y) - \beta \log \frac{\pi(y|x;\theta)}{\pi_{ref}(y|x)} \right]
\end{equation*}
where $ \pi(y|x;\theta) $ is the current policy, $ \pi_{ref}(y|x) $ is the supervised fine-tuned (SFT) policy, and $ \beta $ controls policy shifts.  

Reward modeling requires extensive data labeling. In this paper, we focus on policy optimization with pre-trained reward models, aiming to optimize multiple reward objectives simultaneously.  

\textbf{Multi-Objective Optimization.} Formally, the MOO problem is defined as:
\begin{equation}
    \max_{\theta} (J^{(1)} (\theta), J^{(2)} (\theta), \dots, J^{(N)} (\theta))^\top,
\end{equation}
where $ \theta $ denotes the learnable parameters, $ J^{(i)} $ represents the $ i $-th optimization objective, and the goal is to find a Pareto optimal solution.   
\begin{definition}[Pareto Optimality]
A solution $ \pi^* $ is Pareto optimal if no other solution dominates it, i.e., there does not exist another policy~$ \pi' $ such that:
\begin{itemize}
    \item $ J_i(\pi') \geq J_i(\pi^*) $ for all $ i $.
    \item $ J_j(\pi') > J_j(\pi^*) $ for at least one $ j $.
\end{itemize}
\end{definition}
Since direct vector-form optimization is intractable, MOO is often scalarized into a weighted sum:
\begin{equation}
\label{eq:scalarized_multi_objective}
    \min_{\theta}  \sum_{i=1}^{N} c^{(i)} \mathcal{L}^{(i)} (\theta),
\end{equation}
where $ c^{(i)} $ denotes the weight assigned to each objective $ \mathcal{L}^{(i)} $.  

\textbf{Optimization Challenges.} Solving \cref{eq:scalarized_multi_objective} presents several challenges: i) Balancing conflicting objectives. LLMs often exhibit strong trade-offs between objectives, making simple scalarization ineffective: it can bias solutions toward certain objectives while neglecting others. ii) Weight sensitivity. The choice of weights $ c^{(i)} $ significantly impacts optimization and is often subjective. Poorly chosen weights can lead to suboptimal or undesired solutions. iii) Computational Complexity. Gradient-based multi-objective learning methods generally require computing full gradients for all objectives across all parameters and operate on the gradient with $\mathcal{O}(n^2d)$ complexity (detailed in~\cref{app sec: Complexity Analysis of Multi Objective Optimization Algorithms}). This becomes infeasible at LLM scale due to the high parameter count.  

To address these challenges, we introduce PAMA, a scalable optimization algorithm that ensures convergence to a Pareto stationary point.

\subsection{Noon PPO}
\label{sec:noon_ppo}
We introduce Noon PPO, a variant of PPO~\citep{PPO}, designed to improve stability in MOA. Noon stands for ``No Negative'', as it modifies the advantage to disregard negative values, thereby restricting policy updates to actions with non-negative advantages.
Let $A_t'$ denote the estimated advantage at time step $t$. In Noon PPO, we define the advantage as:
\begin{equation}
A_t = \max\bigl(A_t', 0\bigr).
\label{eq:noon_advantage}
\end{equation}
This adjustment ensures that only actions with a positive advantage contribute to the policy gradient update, effectively ignoring updates that would reduce the probability of suboptimal actions.
As in standard PPO, let $\pi_\theta$ be the current policy parameterized by $\theta$, and let $\pi_{\theta_{\text{ref}}}$ represent the SFT policy. The probability ratio is defined as:
\begin{equation}
u_t(\theta) = \frac{\pi_\theta(a_t \mid s_t)}{\pi_{\theta_{\text{ref}}}(a_t \mid s_t)}.
\label{eq:noon_ratio}
\end{equation}
The clipped surrogate objective in Noon PPO is then given by:
\begin{equation}
\mathcal{L}^{\mathrm{NOON}}(\theta) = {\mathbb{E}}_t \Bigl[
\min\bigl(u_t(\theta)\,A_t,\;
\mathrm{clip}\bigl(u_t(\theta),1-\epsilon,1+\epsilon\bigr)\,A_t\bigr)
\Bigr],
\label{eq:noon_objective}
\end{equation}
where $A_t$ is defined in Equation~\ref{eq:noon_advantage}, and $\epsilon$ is a clipping hyperparameter that limits the deviation between $\pi_\theta$ and $\pi_{\theta_{\text{ref}}}$.

By clipping negative advantages to zero, Noon PPO eliminates unstable gradient fluctuations caused by error-prone or ambiguous training examples. This leads to more predictable convergence, which is particularly beneficial when aligning LLMs with multiple objectives. As we will discuss in ~\cref{sec: PAMA Theoretical Guarantee}, this design plays a crucial role in ensuring the theoretical convergence of PAMA.
\subsection{\label{sec: Solving multi-objective learning problems of LLMs scale}Solving Multi-Objective Optimization at LLM scale}

Optimizing multiple conflicting objectives in LLMs is a challenging task, especially when relying on gradient-based MOO methods~\citep{MGDA, pcgrad, cagrad, ica}. These methods require solving complex gradient aggregation problems, which become computationally infeasible at the scale of modern LLMs. For example, MGDA~\citep{MGDA} formulates the gradient balancing problem as a min-norm optimization, which has a computational cost of $\mathcal{O}(n^2d)$, where $d$ is the model's parameter dimension. Given that $d$ often reaches billions in large-scale models (e.g., 7B parameters), these approaches are prohibitively expensive in both computation and memory, as further discussed in~\cref{app sec: mgda-ub discussion}.

\textbf{Motivation for PAMA.} To overcome these limitations, an efficient and scalable optimization strategy is required. Ideally, such a method should:
\begin{enumerate}
    \item Avoid costly gradient-based operations that scale poorly with model size.
    \item Provide a computationally tractable formulation that remains efficient as the number of objectives grows.
    \item Ensure convergence to a well-defined Pareto stationary point, effectively balancing multiple objectives.
\end{enumerate}
We introduce Pareto Multi-Objective Alignment (PAMA), a novel algorithm specifically designed for large-scale LLM alignment. Instead of directly solving the expensive min-norm optimization, PAMA reformulates the problem into a convex optimization framework with a closed-form solution. This transformation reduces the computational complexity from $\mathcal{O}(n^2d)$ to $\mathcal{O}(n)$, where $n$ is the number of objectives, significantly lowering the computational burden compared to traditional methods.

A key challenge in MOO is determining an appropriate convex combination of gradient directions that balances competing objectives. The conventional approach~\citep{MGDA} relies on solving the min-norm optimization problem:
\begin{equation}
\begin{aligned}
\label{eq: original min norm problem}
\min_{c^{(1)}, \ldots, c^{(N)}} & \left\{ 
\left\| \sum_{i=1}^{N} c^{(i)} \nabla_{\theta} {\mathcal{L}}^{(i)} (\theta) \right\|_2^2 \right.  \left. \text{s.t.} \quad \sum_{i=1}^{N} c^{(i)} = 1, \quad c^{(i)} \geq 0  \quad \forall i 
\right\}
\end{aligned}
\end{equation}
where $\mathcal{L}^{(i)}$ represents the loss for the $i$-th objective, and $c^{(i)}$ is the weight assigned to its gradient contribution. Recent advances~\citep{MGDA} showed that this optimization either results in a KKT stationary point (indicating a Pareto stationary solution) or finds a direction that improves all objectives. However, solving this problem at LLM scale remains intractable due to the high dimensionality of the parameter space.

To mitigate this issue, we derive an upper bound for the min-norm formulation with Noon PPO objectives, which leads to a more efficient optimization approach. Specifically, we show that:
\begin{equation}
\label{eq: upper bound of min norm}
    \begin{aligned}
          \left\| \sum_{i=1}^{N} c^{(i)} \nabla_{\theta} \mathcal{L}^{(i)} (\theta)  \right\|_2^2 &=  \left\| \sum_{i=1}^{N} c^{(i)} \nabla_{\pi} \mathcal{L}^{(i)} (\theta) \nabla_\theta \pi(\theta) \right\|_2^2 =  \left\| \sum_{i=1}^{N} c^{(i)} \frac{1}{\pi_{ref}} I(A^{(i)}) \nabla_\theta \pi(\theta) \right\|_2^2 \\ 
         & \leq  \left\| \sum_{i=1}^{N} c^{(i)}  I(A^{(i)})  \right\|_2^2 \left\| \frac{1}{\pi_{ref}} \nabla_\theta \pi(\theta) \right\|_2^2,
    \end{aligned}
\end{equation}
where 
\begin{equation}
    \begin{aligned}
        I(A) = 
        \begin{cases} 
            0, &  u > 1 + \epsilon \\
            A, &  u \leq 1 + \epsilon 
        \end{cases}, 
    \end{aligned}
\end{equation}
\begin{equation}
    \sum_{i=1}^{N} c^{(i)} = 1, \quad c^{(i)} \geq 0  \quad \forall i,
\end{equation}
and $u=\frac{\pi}{\pi_{ref}}$. For simplicity, we omit the expectation notation, which does not affect the theoretical derivation. The second equation follows from the Noon PPO loss~\cref{eq:noon_objective}, while the final inequality is derived from the Cauchy–Schwarz inequality. This upper bound allows us to reformulate the problem as a more efficient surrogate optimization:
\begin{equation}
\begin{aligned}
\label{eq: our min norm problem}
\min_{c^{(1)}, \ldots, c^{(N)}} & \left\{ 
\left\| \sum_{i=1}^{N} c^{(i)}  I(A^{(i)})  \right\|_2^2  \right.  \left. \text{s.t.} \quad \sum_{i=1}^{N} c^{(i)} = 1, \quad c^{(N)} \geq 0  \quad \forall i 
\right\}.
\end{aligned}
\end{equation}
This formulation admits a closed-form solution, which we derive next.
\begin{theorem}[Optimal Convex Combination for the Min-Norm Problem]
\label{theorem: Optimal Convex Combination for the Scalar Min-Norm Problem}
Let $A^{(1)}, A^{(2)}, \ldots, A^{(N)} \in \mathbb{R}$ be given, and consider the optimization problem
\begin{equation}\label{eq:opt_problem}
\begin{aligned}
\min_{c^{(1)}, \ldots, c^{(N)}} \quad & \left( \sum_{i=1}^{N} c^{(i)} A^{(i)} \right)^2, \\
\text{subject to} \quad & \sum_{i=1}^{N} c^{(i)} = 1, \\
& c^{(i)} \ge 0 \quad \text{for } i = 1,2,\ldots,N.
\end{aligned}
\end{equation}
Then the optimal value of the convex combination,
\begin{equation}
s^* = \sum_{i=1}^{N} c^{(i)} A^{(i)},
\end{equation}
is given by
\begin{equation}
s^* =
\begin{cases}
0, & \text{if } \min_{1\le i\le N} A^{(i)} \le 0 \le \max_{1\le i\le N} A^{(i)}, \\[1mm]
\min_{1\le i\le N} A^{(i)}, & \text{if } A^{(i)} > 0 \text{ for all } i, \\[1mm]
\max_{1\le i\le N} A^{(i)}, & \text{if } A^{(i)} < 0 \text{ for all } i.
\end{cases}
\end{equation}
In other words, $s^*$ is the projection of $0$ onto the interval
\begin{equation}
\left[\min_{1\le i\le N} A^{(i)},\, \max_{1\le i\le N} A^{(i)}\right],
\end{equation}
and the minimum objective value is $(s^*)^2$.
\end{theorem}
The proof is provided in \cref{app: sec: proof of closed form solution for min-norm}.

\textbf{Advantages of PAMA's Reformulation.} Compared to the intractable original optimization problem~(\cref{eq: original min norm problem}), our reformulation provides two key benefits:
\begin{enumerate}
    \item Drastic reduction in computational cost: The term $I(A^{(i)})$ is computed via a simple forward pass, eliminating costly backpropagation.
    \item Analytically solvable optimization: The surrogate problem admits a closed-form solution~(\cref{theorem: Optimal Convex Combination for the Scalar Min-Norm Problem}), ensuring efficiency..
\end{enumerate}

By incorporating this approach with the Noon PPO, we obtain a practical and scalable algorithm for MOA. We summarize PAMA in \cref{app sec: pseudocode}. To illustrate the computational efficiency of our method, consider the magnitude of operations required. Traditional approaches with a complexity of $\mathcal{O}(n^2d)$ result in a computational load of approximately $10^{12}$ operations when $d \approx 10^{10}$ and $n \approx 10^1$. In contrast, our method, operating with $\mathcal{O}(n)$ complexity, requires 10 operations, a very small number. Our approach remains practical even for extremely large-scale problems.

\subsection{Theoretical Guarantee~\label{sec: PAMA Theoretical Guarantee}}

With the reformulated optimization problem in \cref{eq: our min norm problem}, an important question arises: does our approach retain theoretical guarantees? In this section, we establish that under mild conditions, our method converges to a Pareto stationary point, ensuring that no objective can be improved without deteriorating at least one other objective.

First, we formally define the notion of a Pareto stationary point, which serves as a necessary condition for Pareto optimality.
\begin{definition}[Pareto Stationary Point] \label{def: Pareto Stationary Point} 
A parameter vector $\theta$ is said to be satisfying Pareto stationary if there exists a set of weights $\{c^{(i)}\}_{i=1}^N$ satisfying
\begin{equation} 
\sum_{i=1}^{N} c^{(i)} = 1, \quad c^{(i)} \geq 0, \quad \forall i \in \{1,2, \cdots, N\}, \quad \text{and} \quad \sum_{i=1}^{N} c^{(i)} \nabla_{\theta} \mathcal{L}^{(i)} (\theta) = 0. 
\end{equation} 
\end{definition}
Pareto stationary point ensures that no descent direction exists that simultaneously improves all objectives, indicating that the optimization has reached a balanced trade-off among competing objectives.
To establish convergence results, we assume that the loss function exhibits smoothness properties, which are commonly satisfied in deep learning due to gradient-based optimization and regularization.
\begin{definition}[$\kappa$-Lipschitz Continuity]
    Let $(X, d_X)$ and $(Y, d_Y)$ be metric spaces. A function $f: X \to Y$ is said to be $\kappa$-Lipschitz continuous if there exists a constant $\kappa \geq 0$ such that for all $x, y \in X$,
\begin{equation}
    d_Y\bigl(f(x), f(y)\bigr) \leq \kappa d_X(x, y).
\end{equation}
\end{definition}
This property ensures that the function does not change too rapidly, contributing to stability in gradient-based optimization.
\begin{assumption}[Lipschitz Smoothness of the Gradient] \label{assumption:lipschitz} The loss function $\mathcal{L}(\theta)$ has a $\kappa$-Lipschitz continuous gradient, meaning there exists a constant $\kappa >0$ such that for all $\theta, \theta'$
\begin{equation}
\|\nabla_{\theta} \mathcal{L}(\theta) - \nabla_{\theta} {\mathcal{L}}(\theta')\|_2 \leq \kappa \|\theta - \theta'\|_2.
\end{equation}
\end{assumption}
This assumption guarantees that the landscape does not contain abrupt changes, which is critical for convergence guarantees and is empirically observed in RL~\citep{DBLP:conf/iclr/IlyasESTJRM20}.
\begin{assumption}[Bounded Learning Rate] \label{assumption:learningrate} The learning rate $\eta$ satisfies
\begin{equation} 0 \leq \eta \leq \frac{2}{\kappa}. \end{equation} \end{assumption}
This condition ensures stable updates, preventing divergence due to excessively large steps, aligning with standard practices in convex and non-convex optimization.
\begin{assumption}[Bounded Reward]
\label{assumption: bounded reward}
Rewards in RL are typically finite due to practical constraints. Formally, there exists a constant \( R_{\max} > 0 \) such that  
\begin{equation}
    |r(x, y)| \leq R_{\max}, \quad \forall (x, y) \in \mathcal{X} \times \mathcal{Y}.
\end{equation}
\end{assumption}
See~\cref{app sec: discussion about reward} for more discussion.

We now establish the convergence of PAMA.
\begin{restatable}[General Descent Lemma]{lemma}{GeneralDescentLemma}
\label{lemma:GeneralDescent}
Let $f: \mathbb{R}^N \to \mathbb{R}$ be continuously differentiable on an open set containing $x\in\mathbb{R}^N$, and suppose that $\nabla f$ is $\kappa$-Lipschitz continuous, i.e., for all $u,v$ in that set,
\begin{equation}
\|\nabla f(u)-\nabla f(v)\| \le \kappa\|u-v\|.
\end{equation}
Then, for any update direction $g\in\mathbb{R}^N$, one has
\begin{equation}
f(x+g) \le f(x) + \nabla f(x)^\top g + \frac{\kappa}{2}\|g\|^2.
\end{equation}
\end{restatable}
The proof is in~\cref{app sec: proof of general descent}. Using this result, we analyze the gradient descent dynamics of PAMA and show that PAMA converges to a Pareto stationary point. 
\begin{restatable}[Convergence of PAMA]{theorem}{ConvergencePAMA}
\label{theorem: convergence of min-norm upperbound}
Let $ {\mathcal{L}}^{(i)}(\theta) $ be the loss function for task $ i $, where policy is $ \pi(\theta)$. Define the PAMA gradient aggregation:
\begin{equation}
g_o^{(k)} = \sum_{i=1}^N c^{(i)} \nabla_{\pi} {\mathcal{L}}^{(i)}(\theta_k),
\end{equation}
where $ c^{(i)} $ is the solution to 
\begin{equation}
\min_{c^{(1)},\ldots,c^{(N)}} \Big\| g_o \Big\|_2^2, \quad \text{s.t. } \sum_{i=1}^N c^{(i)} = 1, \quad c^{(i)} \geq 0.
\end{equation}
Under \cref{assumption:lipschitz,assumption:learningrate,assumption: bounded reward}, the gradient descent update at timestep $k$:
\begin{equation}
\theta_{k+1} = \theta_k - g_o^{(k)} \eta J 
\end{equation}
ensures 
\begin{equation}
\lim_{k\to\infty} \|\nabla_{\theta} {\mathcal{L}}(\theta_k)\|_2 = 0,
\end{equation}
where $J=\nabla_{\theta_k} \pi(\theta_k)$ and $J \in \mathbb{R}^{ |\theta|\times 1}$.
This shows the update converges to a Pareto stationary point.
\end{restatable}
The proof is provided in~\cref{app sec proof of pama}. \Cref{theorem: convergence of min-norm upperbound} establishes that:
\begin{itemize}
    \item If the optimal value of \cref{eq: our min norm problem} is zero, the aggregated gradient vanishes, indicating that the process has reached a Pareto stationary point.
    \item If the optimal value is nonzero, the gradient provides a valid descent direction for all objectives, ensuring continual improvement toward a Pareto stationary solution.
\end{itemize}
Thus, PAMA guarantees convergence to a balanced trade-off among conflicting objectives, offering a provably convergent and computationally efficient approach to multi-objective alignment for LLMs.

\section{Experiments~\label{sec: exp}}

In this section, we aim to empirically validate whether the theoretical advantages of PAMA are reflected in practical experiments. To this end, we conduct systematic evaluations across different model scales and diverse, potentially conflicting objectives to assess PAMA's effectiveness in multi-objective alignment.

We conduct experiments on three progressively larger language models: GPT-2 (125M), GPT-2 XL (1.5B), and LLaMA-2 (7B), and evaluate PAMA using a range of reward models, including harmlessness, humor, sentiment, and response length. Our implementation is based on the open-source TRL framework~\citep{vonwerra2022trl}. All experiments are conducted on a workstation equipped with an Intel i9-14900K CPU and a single NVIDIA RTX A6000 GPU. Further experimental details are provided in~\cref{app sec: Additional Details Regarding Experiments}, with additional results included in~\cref{app sec: Additional Experimental Results}.
\begin{figure*}[!htbp]
    \centering
    \begin{subfigure}{0.48\linewidth}
        \centering
        \includegraphics[width=\linewidth]{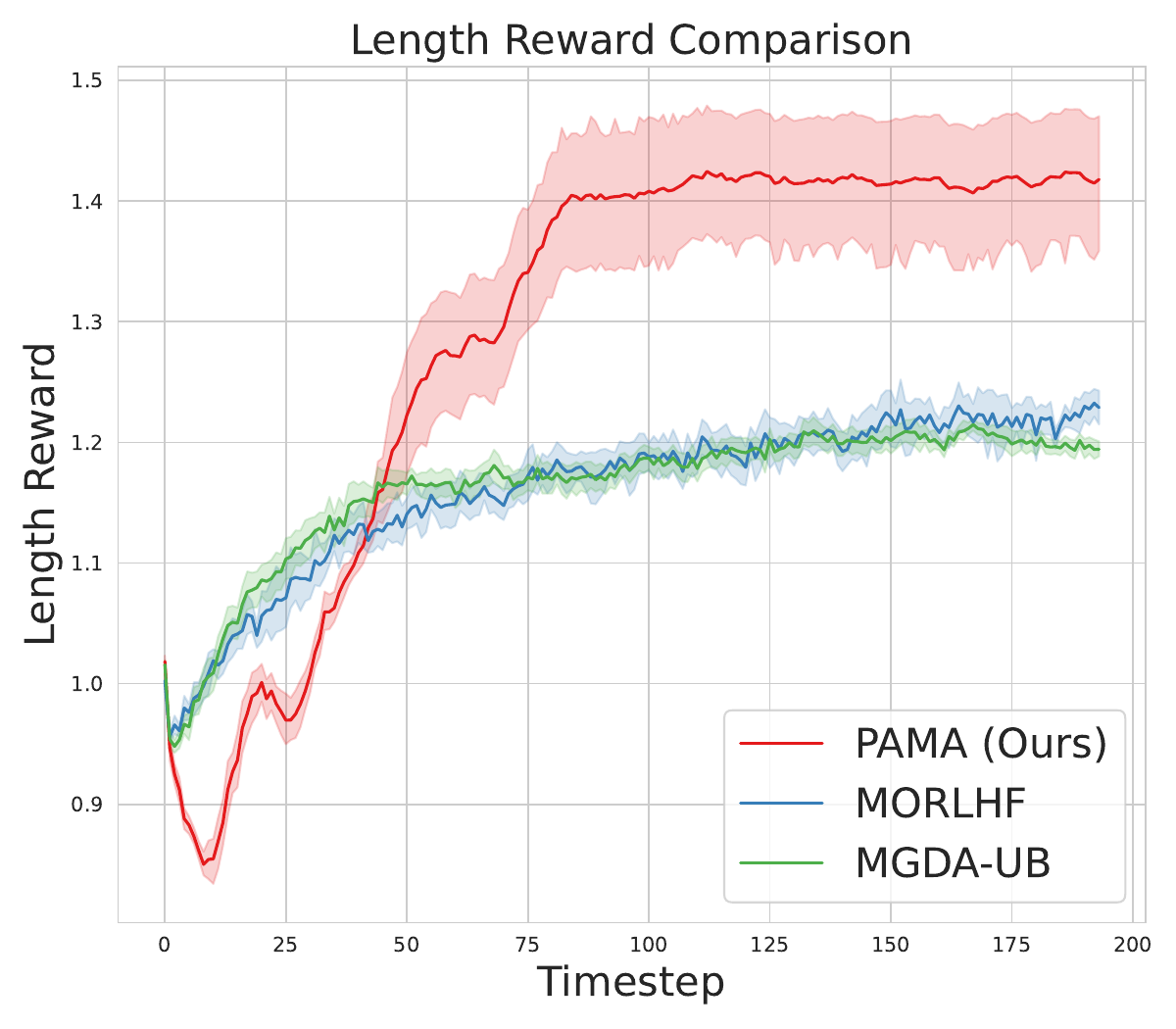}
        \caption{Length}
        \label{fig: gpt2_length_reward}
    \end{subfigure}
    \hfill
    \begin{subfigure}{0.48\linewidth}
        \centering
        \includegraphics[width=\linewidth]{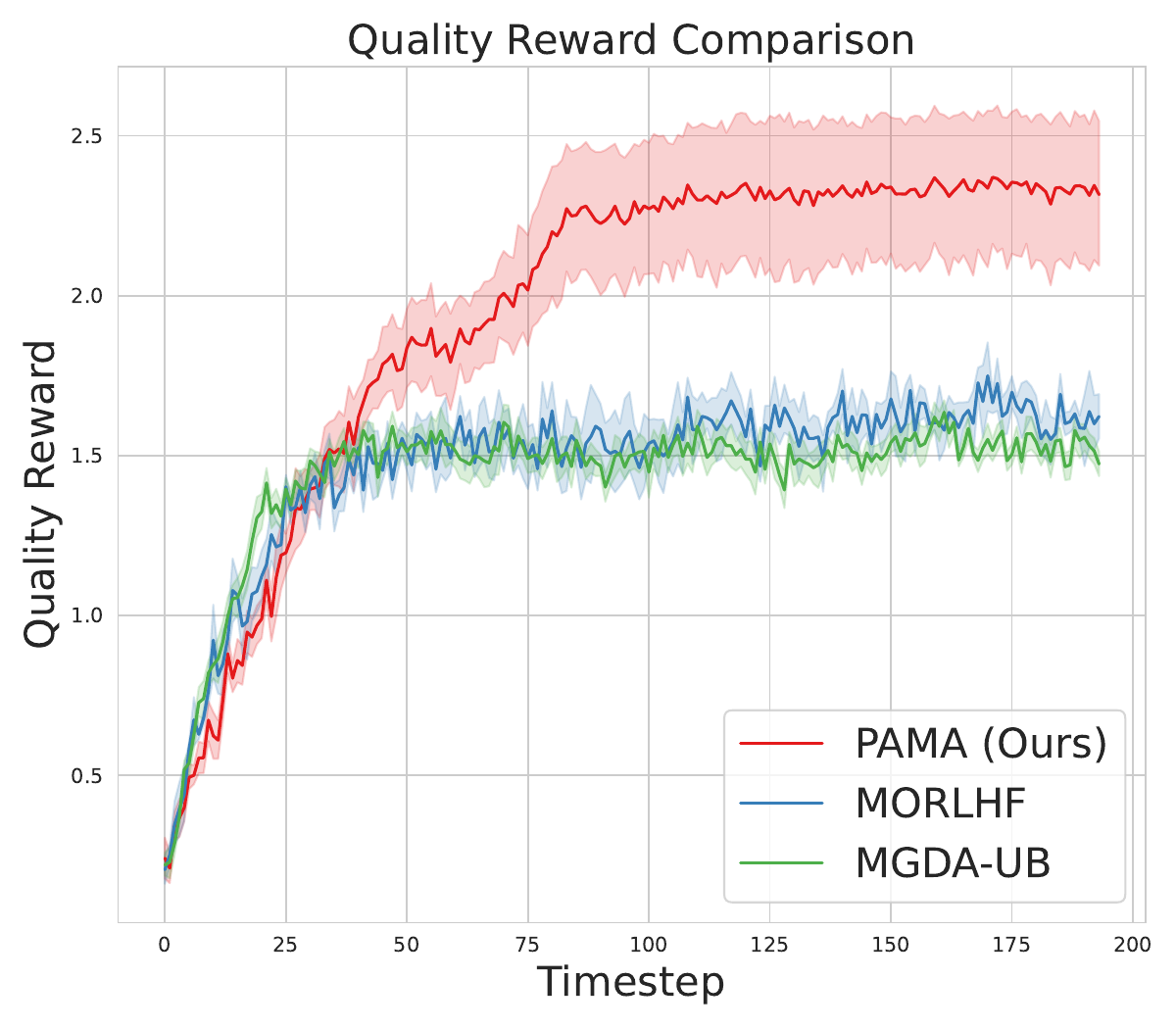}
        \caption{Sentiment}
        \label{fig: gpt2_quality_reward}
    \end{subfigure}
    \caption{Comparison of sentiment and length rewards during training on the IMDb dataset using GPT-2 (125M parameters). PAMA consistently achieves superior performance across both objectives, demonstrating stable optimization. In contrast, MORLHF struggles to balance sentiment and length due to the limitations of the fixed weighted sum approach, while MGDA-UB does not show any advantage over MORLHF. The shaded area represents the standard deviation over eight trials, highlighting the robustness of PAMA.}
    \label{fig: gpt2_length_quality_comparison}
\end{figure*}
\subsection{Normal Model: GPT-2 (125M Parameters)}

In this experiment, we evaluate PAMA on a normal-scale language model, GPT-2 (125M parameters), to assess its effectiveness in optimizing multiple objectives. Specifically, we aim to generate film reviews that are both positive and long, requiring the model to balance sentiment and length objectives.

\textbf{Setup.} We use GPT-2~\citep{gpt1} as the base model and train it on the IMDb dataset\footnote{\url{https://huggingface.co/datasets/stanfordnlp/imdb}}. The objective consists of two reward functions: i) a pretrained sentiment analysis model\footnote{\url{https://huggingface.co/lvwerra/distilbert-imdb}}, where the logit output serves as the reward signal to encourage positive reviews, and ii) a length-based reward that promotes longer responses. Both reward values are structured such that higher scores indicate better performance.

\textbf{Baselines.} We compare PAMA against two widely used baselines: MORLHF, which applies a fixed weighted sum of the objectives, a common but often suboptimal approach for balancing conflicting goals; and MGDA-UB~\citep{mgda_nips}, which leverages the min-norm algorithm to compute gradients that balance multiple objectives dynamically. Further discussion is provided in~\cref{app sec: mgda-ub discussion}.

\textbf{Results.} The training curves in~\cref{fig: gpt2_length_quality_comparison} illustrate the performance of different methods over time. \cref{fig: gpt2_length_reward} shows that PAMA significantly outperforms both baselines in optimizing the length reward. While MORLHF and MGDA-UB exhibit slow and marginal improvements, PAMA achieves a much higher final reward with a stable convergence pattern. \cref{fig: gpt2_quality_reward} further highlights PAMA's advantage in optimizing sentiment, where it reaches a substantially higher reward than the baselines. In contrast, MORLHF stagnates at a lower level, and MGDA-UB shows negative improvement over MORLHF. 
\subsection{Scaling Up: GPT-2 XL (1.5B Parameters)}
\begin{figure*}[!htbp]
    \centering
    \begin{subfigure}{0.48\linewidth}
        \centering
        \includegraphics[width=\linewidth]{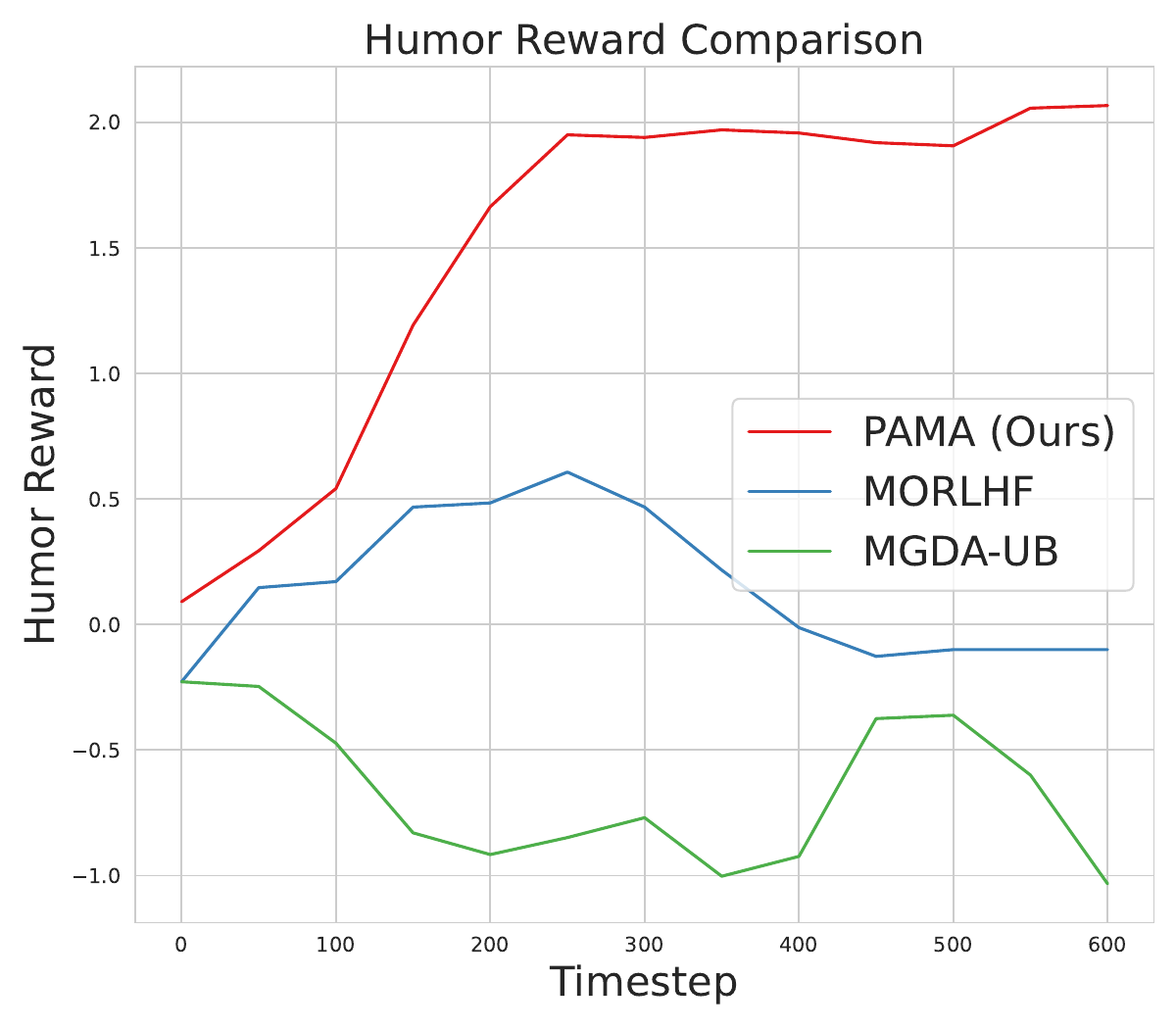}
        \caption{Humor}
        \label{fig: gpt2xl_humor_reward}
    \end{subfigure}
    \hfill
    \begin{subfigure}{0.48\linewidth}
        \centering
        \includegraphics[width=\linewidth]{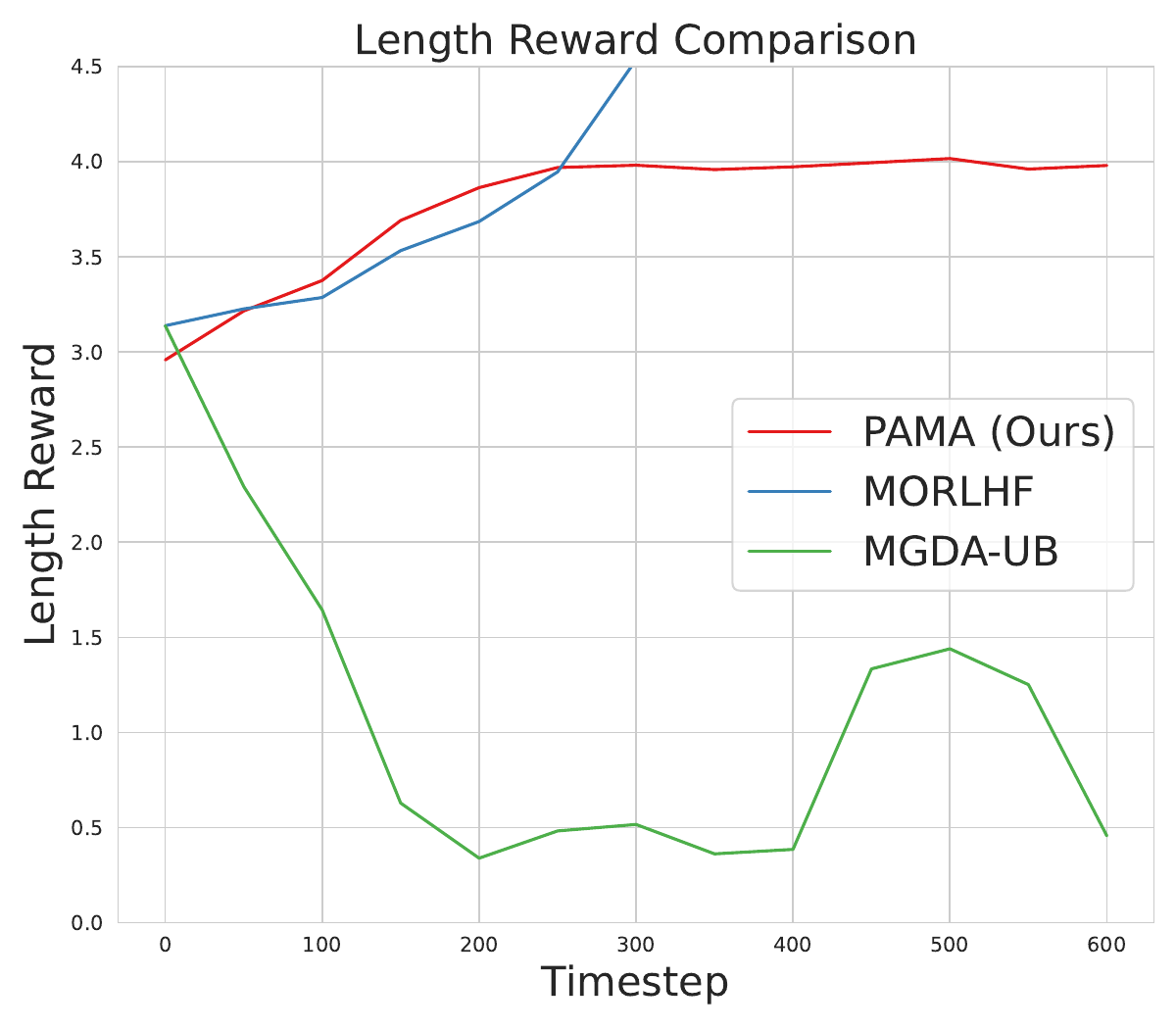}
        \caption{Length}
        \label{fig: gpt2xl_length_reward}
    \end{subfigure}
    \caption{\label{fig: gpt2xl_humor_length_comparison}Comparison of humor and length rewards during training on the HH-RLHF dataset using GPT-2 XL (1.5B parameters). PAMA consistently outperforms the baselines in both objectives, demonstrating stable optimization. While MORLHF fails to significantly improve humor. MGDA-UB struggles in both objectives, showing severe performance degradation. These results highlight the effectiveness of PAMA in multi-objective alignment for LLMs.
    }
\end{figure*}
To evaluate PAMA's scalability and adaptability, we extend our experiments to GPT-2 XL (1.5B parameters), optimizing for both humor and text length.

We train GPT-2 XL on the HH-RLHF dataset~\citep{hh-rlhf} while optimizing two distinct reward signals: i) a humor classifier\footnote{\url{https://huggingface.co/mohameddhiab/humor-no-humor}}, which assigns higher rewards to funnier outputs, and ii) a length-based reward that promotes longer responses. Higher reward values correspond to better performance in both objectives. We compare PAMA against MORLHF and MGDA-UB.

\textbf{Results.} The evaluation results, shown in~\cref{fig: gpt2xl_humor_length_comparison}, illustrate the performance on the test set for humor and length rewards over training timesteps. \cref{fig: gpt2xl_humor_reward} demonstrates that PAMA effectively optimizes humor, steadily increasing its reward and maintaining a high final value. In contrast, MORLHF shows only marginal improvement before plateauing at a lower level, while MGDA-UB fails entirely, with its humor reward even decreasing over time. \cref{fig: gpt2xl_length_reward} shows that both PAMA and MORLHF successfully optimize length, though MORLHF only optimizes length, ignoring humor. MGDA-UB, on the other hand, completely collapses in this setting, with its length reward deteriorating throughout training. These findings reinforce PAMA's robustness in multi-objective alignment, particularly in balancing competing rewards while ensuring stable convergence.
\subsection{Towards Large Language Models: LLaMA-2 7B}

\begin{figure*}[!htbp]
    \centering
    \begin{subfigure}{0.48\linewidth}
        \centering
        \includegraphics[width=\linewidth]{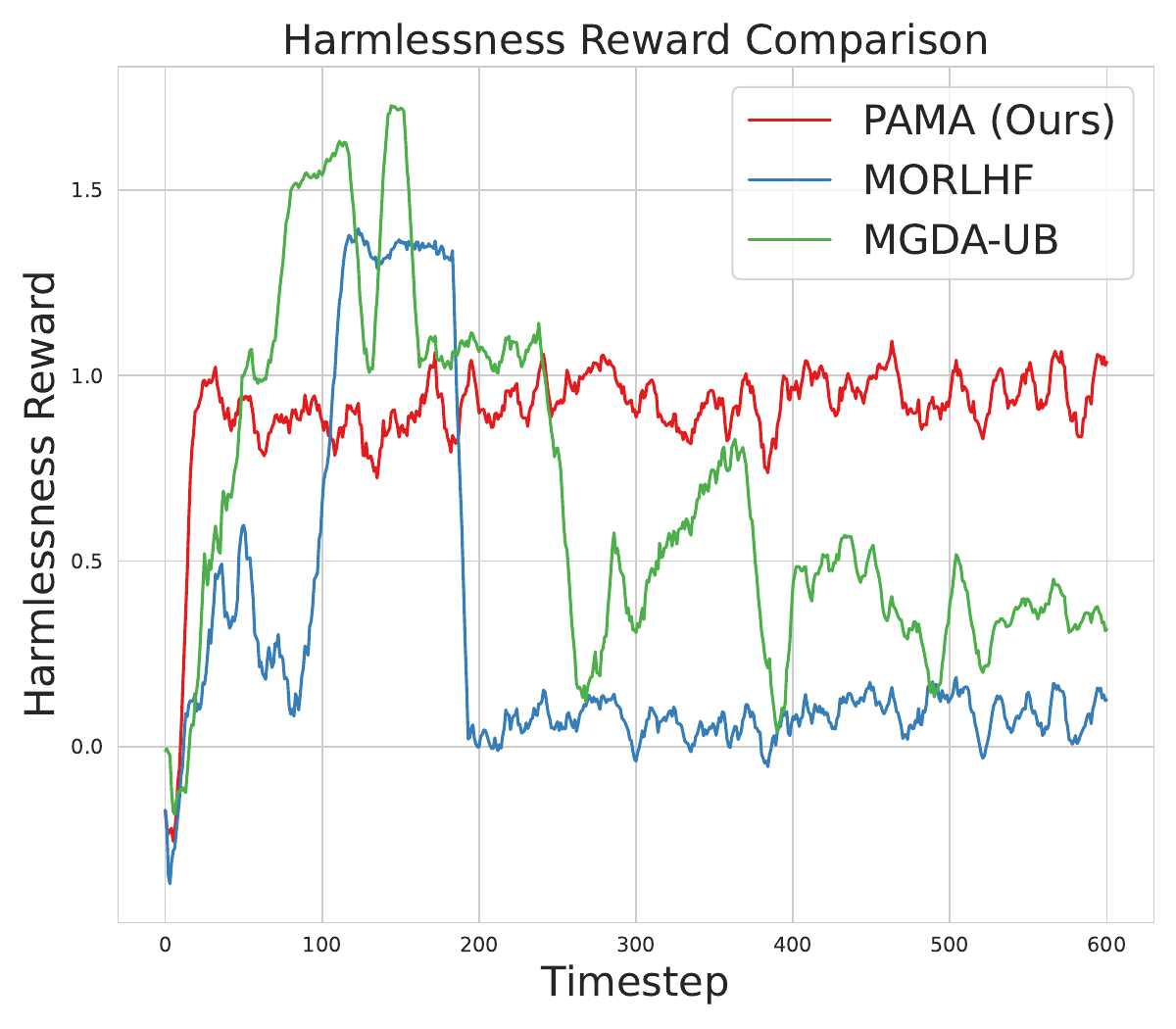}
        \caption{Harmlessness}
        \label{fig: llama2_harmless_reward}
    \end{subfigure}
    \hfill
    \begin{subfigure}{0.48\linewidth}
        \centering
        \includegraphics[width=\linewidth]{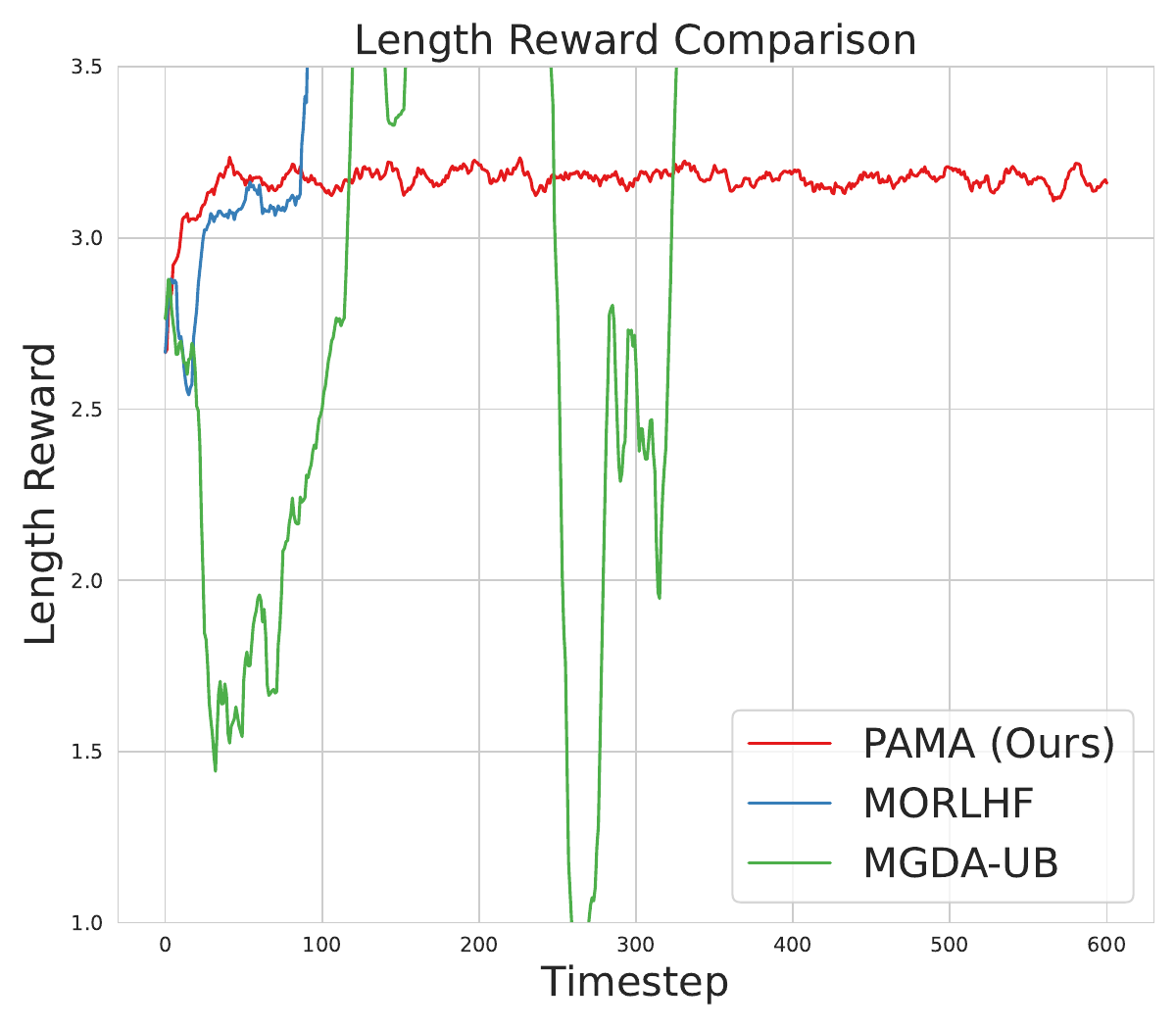}
        \caption{Length}
        \label{fig: llama2_length_reward}
    \end{subfigure}
    \caption{\label{fig: llama2_harmless_length_comparison}Comparison of harmlessness and length rewards during training on the HH-RLHF dataset using LLaMA-2 (7B parameters). PAMA consistently optimizes both objectives while maintaining a stable learning process. In contrast, MGDA-UB and MORLHF struggle with harmlessness optimization, exhibiting significant fluctuations and instability. MGDA-UB, in particular, exhibits pronounced oscillations during training. While MORLHF converges to a lower performance level. These results highlight the robustness of PAMA in aligning large-scale LLMs with multiple objectives.}
\end{figure*}
To assess the scalability of PAMA, we extend our evaluation to a large language model setting using LLaMA-2~\citep{llama2} with 7B parameters. This experiment focuses on aligning the model to generate responses that are both harmless and as long as possible. We utilize the HH-RLHF dataset and measure harmlessness using an open-source reward model\footnote{\url{https://huggingface.co/Ray2333/gpt2-large-harmless-reward_model}}.

\textbf{Results.} The results in~\cref{fig: llama2_harmless_length_comparison} demonstrate PAMA's effectiveness in large-scale multi-objective alignment. As shown in~\cref{fig: llama2_harmless_reward}, PAMA achieves a stable increase in harmlessness reward, while MORLHF and MGDA-UB suffer from instability and fluctuations. MGDA-UB, in particular, exhibits pronounced training oscillations, failing to maintain a high harmlessness score, whereas MORLHF stabilizes at a lower reward level. Similarly,~\cref{fig: llama2_length_reward} illustrates that PAMA maintains strong performance in length optimization, achieving stable convergence. In contrast, MGDA-UB experiences erratic fluctuations, and MORLHF fails to sustain meaningful progress. These findings reinforce PAMA's theoretical advantages, demonstrating its ability to effectively balance competing objectives in large-scale LLM alignment.

\subsection{Discussion}
Our experimental results confirm that the theoretical advantages of PAMA are consistently realized in practice. Across various model size (ranging from 125M to 7B) and objective settings, PAMA demonstrates superior stability and optimization performance, significantly outperforming existing baseline methods.
MORLHF, which relies on a weighted sum of objectives, struggles to balance competing rewards due to its fixed weight assignments, often leading to suboptimal trade-offs. MGDA-UB, while employing dynamic gradient balancing, can exhibit training instability and, in some cases, underperform compared to MORLHF.
These findings highlight PAMA's robustness in achieving stable and well-balanced optimization across different model scales and reward settings, making it a reliable and scalable solution for multi-objective alignment in large language models.

\section{Related Work~\label{sec: related work}}

\textbf{Multi-Objective Optimization} is a fundamental problem in RL and deep learning, where multiple conflicting objectives must be simultaneously optimized, because improving one often leads to the degradation of another. Classical MOO techniques aim to find Pareto-optimal solutions. Among them, simple linearization methods with fixed weights often fail to effectively balance competing objectives~\citep{boyd2004convex}. A more general approach is Pareto-based optimization, which seeks to optimize all objectives simultaneously while maintaining trade-offs. Gradient-based MOO methods, e.g. MGDA~\citep{MGDA}, formulate a common descent direction for all objectives, ensuring simultaneous progress. However, despite their theoretical appeal, these approaches, along with related methods like PCGrad~\citep{pcgrad} and CAGrad~\citep{cagrad}, suffer from computational inefficiencies in high-dimensional parameter spaces, particularly in deep learning. The prohibitive cost of computing and aggregating gradients at LLM scale motivate the development of scalable alternatives, such as our proposed method, PAMA.

\textbf{MORL} extends RL to settings where an agent must learn policies that balance multiple reward functions. Standard MORL approaches include linear scalarization~\citep{morl-survey}, Envelope Q-Learning~\citep{envelopq-learning}, and Pareto Q-learning~\citep{pareto-q-learning}, as well as several recent extensions~\citep{Pareto-Conditioned-Networks,alegre2023sample,DBLP:conf/iclr/LuHY23,DBLP:journals/jair/FeltenTD24,beer,peer}. These methods are widely used in applications that require trade-offs between competing objectives~\citep{mo-gym}. However, their extension to large-scale neural networks, particularly LLMs, remains an open challenge due to computational constraints and the difficulty of balancing conflicting reward signals. A further discussion is provided in~\cref{app sec: mgda-ub discussion,app sec: Complexity Analysis of Multi Objective Optimization Algorithms}.

\textbf{MOO for LLMs.} Applying MOO to LLMs presents additional challenges due to their high-dimensional parameter space and the inherent conflicts between objectives such as fluency, factual accuracy, and safety. Existing MOO techniques often become impractical for LLMs due to the prohibitive cost of computing gradients for each objective. For example, MGDA-UB~\citep{mgda_nips} is proposed as an efficient approximation method, though its behavior on large-scale models can be unstable in practice, as observed in our experiments. Independent Component Alignment (ICA)~\citep{ica} has been explored in multi-task learning for vision models, but its reliance on singular value decomposition introduces numerical instability, particularly when applied to \texttt{float16} or \texttt{bfloat16} formats used in LLM training. A notable recent approach is MOC~\citep{he2025moc}, which trains an LLM as a meta-policy to generate responses aligned with user-defined preferences along the Pareto front. While promising, such approaches still face scalability and optimization challenges when applied to billion-parameter models.

Our approach, PAMA, distinguishes itself from previous methods by: i) Achieving computational efficiency comparable to single-objective RLHF methods, making it scalable to large models. ii) Providing theoretical guarantees of convergence, ensuring stable and reliable optimization. iii) Directly enabling multi-objective alignment in LLMs without relying on computationally expensive gradient manipulation techniques.
By addressing both theoretical and practical limitations of existing methods, PAMA establishes a scalable and principled solution for aligning LLMs with multiple human values.

\section{Conclusion\label{sec: conclusion}}
In this paper, we introduced Pareto Multi-Objective Alignment, a computationally efficient and theoretically grounded algorithm designed to align large language models across multiple, potentially conflicting objectives. By transforming the inherently complex multi-objective reinforcement learning from human feedback problem into a convex optimization framework, PAMA significantly reduces computational complexity, from an impractical $\mathcal{O}(n^2d)$ to $\mathcal{O}(n)$, where $d$ is the number of parameters (billions for LLMs) and $n$ is the number of objectives. This efficiency enables practical multi-objective optimization even for billion-parameter models, expanding the applicability of LLMs across diverse real-world tasks.
From a theoretical perspective, we provided rigorous proofs demonstrating that PAMA converges to Pareto stationary points. The empirical results further substantiate that PAMA not only exhibits theoretical superiority but also achieves stable and efficient multi-objective alignment in real-world applications. By successfully translating its methodological advantages into tangible performance improvements, PAMA provides a computationally efficient and theoretically grounded solution for multi-objective alignment for LLMs.
In summary, PAMA not only addresses a critical gap in current multi-objective alignment methodologies but also offers a scalable, principled, and computationally viable solution for aligning LLMs with multiple human values. By establishing a strong foundation for efficient multi-objective optimization, PAMA paves the way for more adaptable, responsive, and socially aligned AI systems.

\begin{credits}
\subsubsection{\ackname} This research was supported by the German Federal Ministry of Research, Technology and Space under Grant Number 16KISK035.

\end{credits}

\bibliographystyle{splncs04nat}
\bibliography{sample}

\appendix

\tableofcontents
\newpage
\section{\label{app: sec: proof of closed form solution for min-norm}Proof of \cref{theorem: Optimal Convex Combination for the Scalar Min-Norm Problem}}

\begin{proof}
Define 
\begin{equation}
s = \sum_{i=1}^{N} c^{(i)} A^{(i)}.
\end{equation}
Since the coefficient vector 
\begin{equation}
(c^{(1)}, \ldots, c^{(N)}) 
\end{equation}
lies in the standard simplex
\begin{equation}
\Delta = \Bigl\{ (c^{(1)}, \ldots, c^{(N)}) \in \mathbb{R}^N : \sum_{i=1}^{N} c^{(i)} = 1,\; c^{(i)} \ge 0 \text{ for all } i \Bigr\},
\end{equation}
the set of all possible values of $s$ is exactly the convex hull of $\{A^{(1)}, A^{(2)}, \ldots, A^{(N)}\}$:
\begin{equation}
\mathcal{S} = \left\{ s : s = \sum_{i=1}^{N} c^{(i)} A^{(i)},\; (c^{(1)}, \ldots, c^{(N)}) \in \Delta \right\} = \left[\min_{1\le i\le N} A^{(i)},\, \max_{1\le i\le N} A^{(i)}\right].
\end{equation}
Thus, minimizing 
\begin{equation}
\left( \sum_{i=1}^{N} c^{(i)} A^{(i)} \right)^2 = s^2
\end{equation}
over $\Delta$ is equivalent to choosing $s\in\mathcal{S}$ that minimizes $s^2$.

We now consider two cases.

\textbf{Case 1:} $\min_{1\le i\le N} A^{(i)} \le 0 \le \max_{1\le i\le N} A^{(i)}$.
Since $0 \in \mathcal{S}$, there exists a feasible $(c^{(1)}, \ldots, c^{(N)})$ such that 
\begin{equation}
s = 0.
\end{equation}
Thus, the minimum value of $s^2$ is $0$, and the optimal convex combination is $s^* = 0$.

\textbf{Case 2:} Either $A^{(i)} > 0$ for all $i$ or $A^{(i)} < 0$ for all $i$.
Without loss of generality, assume that $A^{(i)} > 0$ for all $i$. Then 
\begin{equation}
\mathcal{S} = \left[\min_{1\le i\le N} A^{(i)},\, \max_{1\le i\le N} A^{(i)}\right] \subset (0,\infty).
\end{equation}
Since the function $f(s)=s^2$ is strictly increasing for $s>0$, the minimum is achieved at the smallest possible $s$, namely,
\begin{equation}
s^* = \min_{1\le i\le N} A^{(i)}.
\end{equation}
A similar argument applies if $A^{(i)} < 0$ for all $i$, in which case 
\begin{equation}
s^* = \max_{1\le i\le N} A^{(i)}.
\end{equation}

For a complete derivation, we now use the method of Lagrange multipliers. We consider the following optimization problem:
\begin{equation}
\min_{\{c^{(i)}\}} \quad \biggl(\sum_{i=1}^N c^{(i)} \, A^{(i)}\biggr)^2
\end{equation}
subject to
\begin{equation}
\sum_{i=1}^N c^{(i)} = 1 
\quad \text{and} \quad 
c^{(i)} \ge 0 \quad \text{for all } i = 1,\dots,N.
\end{equation}
We define 
\begin{equation}
s = \sum_{i=1}^N c^{(i)} \, A^{(i)}.
\end{equation}
Because each feasible vector $(c^{(1)},\dots,c^{(N)})$ lies in the standard simplex, $s$ must lie in 
\begin{equation}
[\min_i A^{(i)}, \; \max_i A^{(i)}].
\end{equation}
Thus, minimizing $s^2$ is equivalent to projecting $0$ onto the interval 
$[\min_i A^{(i)}, \; \max_i A^{(i)}]$.

To apply the KKT conditions, we rewrite the constraints in a standard form. First, let
\begin{equation}
g(\mathbf{c}) = \sum_{i=1}^N c^{(i)} - 1 = 0,
\end{equation}
and for each $i$ define the inequality constraint
\begin{equation}
h^{(i)}(\mathbf{c}) = -\,c^{(i)} \le 0,
\end{equation}
which enforces $c^{(i)} \ge 0$. The Lagrangian function then becomes
\begin{equation}
\mathcal{L}(\mathbf{c}, \lambda, \{\mu^{(i)}\}) \;=\;
\bigl(\textstyle{\sum_{i=1}^N c^{(i)} A^{(i)}}\bigr)^2
\;+\;
\lambda \Bigl(\textstyle{\sum_{i=1}^N c^{(i)}} - 1\Bigr)
\;+\;
\sum_{i=1}^N \mu^{(i)} \bigl(-\,c^{(i)}\bigr),
\label{eq:Lagrangian}
\end{equation}
where $\lambda \in \mathbb{R}$ is the Lagrange multiplier for the equality constraint and $\mu^{(i)} \ge 0$ are multipliers for the inequality constraints.

Consider the stationarity condition, the gradient of $\mathcal{L}$ with respect to $c^{(i)}$ must vanish at optimality. Note that
\begin{equation}
\frac{\partial}{\partial c^{(i)}} 
\Bigl(\sum_{j=1}^N c^{(j)} A^{(j)}\Bigr)^2 
\;=\;
2\,s\,A^{(i)},
\end{equation}
\begin{equation}
\frac{\partial}{\partial c^{(i)}} 
\Bigl(\sum_{j=1}^N c^{(j)} - 1\Bigr) 
\;=\;
1,
\end{equation}
\begin{equation}
\frac{\partial}{\partial c^{(i)}} \bigl(-\,c^{(k)}\bigr) 
\;=\;
-\,\mathbf{1}_{\{k=i\}}.
\end{equation}
Hence, from \eqref{eq:Lagrangian}, stationarity for each $i$ is:
\begin{equation}
2\,s\,A^{(i)} \;+\; \lambda \;+\; \sum_{k=1}^N \mu^{(k)} \bigl(-\,\mathbf{1}_{\{k=i\}}\bigr) \;=\; 0,
\label{eq:stationary-raw}
\end{equation}
which simplifies to
\begin{equation}
2\,s\,A^{(i)} + \lambda - \mu^{(i)} = 0.
\label{eq:stationary-simplified}
\end{equation}
The KKT conditions also require:
\begin{equation}
\sum_{i=1}^N c^{(i)} = 1, 
\quad
c^{(i)} \ge 0, 
\quad 
\mu^{(i)} \ge 0,
\end{equation}
and complementary slackness:
\begin{equation}
\mu^{(i)} \, \bigl(-\,c^{(i)}\bigr) = 0 \quad \Longrightarrow \quad \mu^{(i)} c^{(i)} = 0.
\end{equation}
That is, whenever $c^{(i)} > 0$, we must have $\mu^{(i)} = 0$. 

\textbf{Case 1: More than one positive $c^{(i)}$.}\\
If there exist two indices $i$ and $j$ for which $c^{(i)}>0$ and $c^{(j)}>0$, then from 
\begin{equation}
2\,s\,A^{(i)} + \lambda = 0
\quad \text{and}\quad
2\,s\,A^{(j)} + \lambda = 0,
\end{equation}
we obtain $2\,s\,A^{(i)} = 2\,s\,A^{(j)}$. If $A^{(i)} \neq A^{(j)}$, it forces $s = 0$. Thus, mixing two distinct $A^{(i)}$ values can only be optimal if $s=0$, which requires $0$ to lie in the interval $[\min_i A^{(i)},\,\max_i A^{(i)}]$.

\textbf{Case 2: All $A^{(i)} > 0$.}\\
If $A^{(i)}>0$ for all $i$, then $0$ cannot lie in $[\min_i A^{(i)}, \max_i A^{(i)}]$. Since $s^2$ is strictly increasing for $s>0$, the minimizing $s$ is $\min_i A^{(i)}$. The solution is then a single vertex of the simplex (i.e., $c^{(k)}=1$ for the $k$ where $A^{(k)}$ is smallest).

\textbf{Case 3: All $A^{(i)} < 0$.}\\
Similarly, if $A^{(i)}<0$ for all $i$, then minimizing $s^2$ over negative values forces $s=\max_i A^{(i)}$ (the least negative number), again achieved by selecting $c^{(k)}=1$ at that particular $k$.

All cases confirm the geometric picture: the optimal point is the projection of $0$ onto the interval $[\min_i A^{(i)},\,\max_i A^{(i)}]$. Thus, the optimal objective value is $(s^*)^2$, where
\begin{equation}
s^* \;=\;
\begin{cases}
0, & \text{if } \min_{i} A^{(i)} \le 0 \le \max_{i} A^{(i)},\\[2pt]
\min_i A^{(i)}, & \text{if } A^{(i)} > 0 \text{ for all } i,\\[2pt]
\max_i A^{(i)}, & \text{if } A^{(i)} < 0 \text{ for all } i.
\end{cases}
\end{equation}
\qed
\end{proof}

\newpage
\section{\label{app sec: proof of general descent}Proof of~\cref{lemma:GeneralDescent}}

\begin{proof}
Define the function $\phi:[0,1]\to\mathbb{R}$ by
\begin{equation}
\phi(t) = f\bigl(x+tg\bigr).
\end{equation}
Clearly, $\phi(0)=f(x)$ and $\phi(1)=f(x+g)$. We now compute the derivative of $\phi$ with respect to $t$. By the chain rule,
\begin{equation}
\frac{d}{dt}\phi(t) = \frac{d}{dt} f\bigl(x+tg\bigr)
=\nabla f\bigl(x+tg\bigr)^\top \frac{d}{dt}\bigl(x+tg\bigr).
\end{equation}
Since
\begin{equation}
\frac{d}{dt}\bigl(x+tg\bigr) = g,
\end{equation}
it follows that
\begin{equation}
\phi'(t)= \nabla f\bigl(x+tg\bigr)^\top g.
\end{equation}
By the Fundamental Theorem of Calculus,
\begin{equation}
f(x+g)=\phi(1)=\phi(0)+\int_0^1 \phi'(t)dt
=f(x)+\int_0^1 \nabla f\bigl(x+tg\bigr)^\top gdt.
\end{equation}
Now, add and subtract $\nabla f(x)$ inside the integral:
\begin{equation}
f(x+g)=f(x)+\nabla f(x)^\top g + \int_0^1 \Bigl[\nabla f\bigl(x+tg\bigr)-\nabla f(x)\Bigr]^\top gdt.
\end{equation}
Since $\nabla f$ is $\kappa$-Lipschitz continuous, for any $t\in[0,1]$ we have
\begin{equation}
\|\nabla f\bigl(x+tg\bigr)-\nabla f(x)\| \le \kappa\|tg\| = \kappa t\|g\|.
\end{equation}
Then, by the Cauchy–Schwarz inequality,
\begin{equation}
\Bigl|\Bigl[\nabla f\bigl(x+tg\bigr)-\nabla f(x)\Bigr]^\top g\Bigr|
\le \|\nabla f\bigl(x+tg\bigr)-\nabla f(x)\|\|g\|
\le \kappa t\|g\|^2.
\end{equation}
Integrating over $t\in[0,1]$, we obtain
\begin{equation}
\left|\int_0^1 \Bigl[\nabla f\bigl(x+tg\bigr)-\nabla f(x)\Bigr]^\top gdt\right|
\le \kappa\|g\|^2 \int_0^1 tdt
= \frac{\kappa}{2}\|g\|^2.
\end{equation}
Thus, combining the above, we conclude that
\begin{equation}
f(x+g) \le f(x)+\nabla f(x)^\top g + \frac{\kappa}{2}\|g\|^2,
\end{equation}
which completes the proof.
\qed
\end{proof}

\newpage
\section{\label{app sec: discussion about reward}Further Discussion About Rewards}

In reinforcement learning, it is generally assumed that the reward function $r(x, y)$ is finite in practical applications, as rewards are typically defined within a bounded range due to physical constraints, human preference modeling, or numerical stability considerations. Formally, we assume there exists a finite constant $R_{\max} > 0$ such that  
    \begin{equation}
        |r(x, y)| \leq R_{\max}, \quad \forall (x, y) \in \mathcal{X} \times \mathcal{Y}.
    \end{equation}

From this assumption, it follows directly that the loss function of Noon PPO remains finite.

\textbf{Bounded Advantage Estimation:} The generalized advantage estimation (GAE)~\citep{PPO} is defined as  
    \begin{equation}
    \label{app eq: ppo advantage}
        A_t = \sum_{l=0}^{\infty} (\gamma \lambda)^l \delta_{t+l},
    \end{equation}
    where $\delta_t = r_t + \gamma V(s_{t+1}) - V(s_t)$ is the temporal difference (TD) residual, and $V(s)$ is the value function. Given that the reward $r_t$ is bounded by $R_{\max}$ and the value function $V(s)$ is finite, it follows that $|\delta_t|$ is bounded. Since $\gamma, \lambda \in [0,1]$, the infinite summation defining $A_t$ converges, ensuring that there exists a constant $A_{\max} > 0$ such that  
    \begin{equation}
        |A_t| \leq A_{\max}, \quad \forall t.
    \end{equation}

\textbf{Bounded Noon PPO Loss Function:} The Noon PPO loss function is given by  
\begin{equation}
    L_{\text{PPO}}(\theta) = \mathbb{E}_{(x, y) \sim D} \left[ \min \left( u(\theta) A, \text{clip}(u(\theta), 1-\epsilon, 1+\epsilon) A \right) \right],
\end{equation}
where $u(\theta) = \frac{\pi_{\theta}(y | x)}{\pi_{\theta_{\text{ref}}}(y | x)}$ is the probability ratio, $A$ is in~\cref{eq:noon_advantage}, and $\epsilon$ is the clipping parameter. Since $A_t$ is bounded by $A_{\max}$, and $u(\theta)$ is constrained within the interval $[1-\epsilon, 1+\epsilon]$ due to clipping, the Noon PPO loss remains finite. That is, there exists a constant $C > 0$ such that  
\begin{equation}
    |L_{\text{Noon PPO}}(\theta)| \leq C, \quad \forall \theta.
\end{equation}

These boundedness ensure that the Noon PPO loss function remains well-defined and does not diverge, providing theoretical stability guarantees for multi-objective optimization.

\newpage
\section{\label{app sec proof of pama}Proof of \cref{theorem: convergence of min-norm upperbound}}

Before presenting the proof, the authors note that the best theoretical guarantee for current gradient-based MOO methods is limited to convergence to Pareto stationary points~\citep{pcgrad,cagrad,MGDA,ica}.

\begin{proof}
PAMA aims to get the co-efficients $c^{(i)}$ by solving:
\begin{equation}
\label{eq: min-norm problem in the proof}
\min_{c^{(1)},\ldots,c^{(N)}} \Biggl\{ \Bigl\|\sum_{i=1}^N c^{(i)} \nabla_{\pi} {\mathcal{L}}^{(i)}(\theta) \Bigr\|_2^2 \;\Bigg|\; \sum_{i=1}^N c^{(i)} = 1,\; c^{(i)} \ge 0\; \forall t \Biggr\}.
\end{equation}
Let the optimal solution be denoted as:
\begin{equation}
\label{eq: min-norm solution}
g_o^{(k)} = \sum_{i=1}^N c^{(i)} \nabla_{\pi} \mathcal{L}^{(i)}(\theta_k),
\end{equation}
where $g_o^{(k)} \in \mathbb{R}$. Without loss of generality, we consider one of the loss functions $\mathcal{L}(\theta)$. The following procedures hold for any loss function in the optimization. 
The gradient descent update rule of parameters $\theta$ is:
\begin{equation}
\theta_{k+1} = \theta_k - \eta [G_{k}],
\end{equation}
where $k$ denotes the timestep and $\eta$ is the learning rate. In PAMA, the gradient of the loss function $\mathcal{L}$ with respect to $\theta$ is:
\begin{equation}
G_{k} = \sum_{t=1}^{T} c^{(t)} \nabla_{\theta_k} \mathcal{L}^{(i)}(\theta_k) = \underbrace{ \sum_{t=1}^{T} c^{(t)}  \nabla_{\pi} \mathcal{L}^{(i)}(\theta_k)}_{g_o^{(k)}} \nabla_\theta \pi = J g_o^{(k)}.
\end{equation}
By the $\kappa$-Lipschitz continuity of the gradient (\cref{assumption:lipschitz}) and \cref{lemma:GeneralDescent}, we have:
\begin{equation}
\label{eq:after-plug}
\mathcal{L}(\theta_{k+1})
\le
\mathcal{L}(\theta_k) +
\bigl\langle
   \nabla_{\theta}\mathcal{L}(\theta_k),-\eta g_o^{(k)}J
\bigr\rangle + \frac{\kappa}{2}\eta^2\|g_o^{(k)}J\|_2^{2}.
\end{equation}
Define the decrease in the loss function as
\begin{equation}
    \Delta_{k} \triangleq \mathcal{L}(\theta_k)-\mathcal{L}(\theta_{k+1}).
\end{equation}

Case i): $g_o^{k}=0$:
We have 
\begin{equation}
\begin{aligned}
\mathcal{L}(\theta_{k+1})
& \leq \mathcal{L}(\theta_k) +
\bigl\langle \nabla_{\theta}\mathcal{L}(\theta_k),-\eta 0 J
\bigr\rangle + \frac{\kappa}{2}\eta^2\|0 J\|_2^{2}  \\
&= \mathcal{L}(\theta_k),
\end{aligned}
\end{equation}
i.e., $\mathcal{L}(\theta_{k+1}) \leq \mathcal{L}(\theta_k)$.

Case ii): $g_o^{k}=0$:
Using $\nabla_{\theta} \mathcal{L}(\theta_k) = \nabla_\pi \mathcal{L} \nabla_\theta \pi $, we obtain:

\begin{equation}
\begin{aligned}
    \mathcal{L}(\theta_k) - \mathcal{L}(\theta_{k+1}) & \geq \bigl\langle
   \nabla_{\theta}\mathcal{L}(\theta_k),\eta g_o^{(k)}J
\bigr\rangle - \frac{\kappa}{2}\eta^2\|g_o^{(k)}J\|_2^{2} \\
& = \bigl\langle
   \nabla_{\theta}\mathcal{L}(\theta_k),\eta g_o^{(k)}J
\bigr\rangle - \frac{\kappa}{2}\eta^2 {g_o^{(k)}}^2 \langle J, J \rangle \\
& = \left \langle \left( \nabla_\pi \mathcal{L}  \eta g_o^{(k)} - \frac{\kappa}{2}\eta^2 {g_o^{(k)}}^2\right)   J , J \right \rangle \\
&= \left( \eta g_o^{(k)} \nabla_\pi \mathcal{L}   - \frac{\kappa}{2}\eta^2 {g_o^{(k)}}^2\right) \| J\|_2^2 \\
& =  \eta g_o^{(k)}  \left(  \frac{1}{\pi_{ref}} I(A)   - \frac{\kappa}{2}\eta {g_o^{(k)}}\right) \| J\|_2^2 \\
& \geq \eta g_o^{(k)}  \left( I(A)   - \frac{\kappa}{2}\eta {g_o^{(k)}}\right) \| J\|_2^2 \\
& \geq \eta g_o^{(k)}  \left( I(A)   - {g_o^{(k)}}\right) \| J\|_2^2 \\
& = \frac{\eta}{g_o^{(k)}}   \left( I(A)   - {g_o^{(k)}}\right) \|g_o^{(k)} J \|_2^2 \\
&= \frac{\eta}{g_o^{(k)}}    \left( I(A)   - {g_o^{(k)}}\right) \|G_k \|_2^2 \\
& \geq 0.
\end{aligned}
\end{equation}
The second inequality holds by $\pi \in (0,1)$. The third inequality holds by \cref{assumption:learningrate}. The last inequality holds by \cref{eq:noon_advantage,eq: min-norm problem in the proof,eq: min-norm solution}.
To handle both cases, define the auxiliary function
\begin{equation}
    f(g_o^{(k)}) =
\begin{cases}
\displaystyle \frac{\eta}{g_o^{(k)}} \Bigl(I(A)-g_o^{(k)}\Bigr) \|G_k\|_2^2, & \text{if } g_o^{(k)}>0, \\[1ex]
\|G_k\|_2^2, & \text{if } g_o^{(k)}=0.
\end{cases}
\end{equation}
Then, for every iteration $k$ we have
\begin{equation}
\Delta_{k} \ge f(g_o^{(k)}).
\end{equation}
Summing over all iterations yields
\begin{equation}
\sum_{k=0}^{\infty} \Delta_k = \mathcal{L}(\theta_0) - \mathcal{L}(\theta_{\infty}) \geq  \sum_{\{k: \, g_o^{(k)}>0\}} \frac{\eta}{g_o^{(k)}} \Bigl( I(A) - g_o^{(k)} \Bigr) \|G_k\|_2^2.
\end{equation}
Since the loss $\mathcal{L}(\theta)$ is bounded, the left-hand side is finite, and hence
\begin{equation}
\sum_{k=0}^{\infty} \|G_k\|_2^2 < \infty.
\end{equation}
This implies that
\begin{equation}
\lim_{k\to\infty} \|G_k\|_2^2 = 0.
\end{equation}
Recalling that $G_k = J\,g_o^{(k)}$, it follows that
\begin{equation}
\lim_{k\to\infty} g_o^{(k)} = 0.
\end{equation}
Since
\begin{equation}
g_o^{(k)} = \sum_{i=1}^{N} c^{(i)} \nabla_{\pi} \mathcal{L}^{(i)}(\theta_k),
\end{equation}
with $c^{(i)}\geq 0$ and $\sum_{i=1}^{N} c^{(i)} = 1$, in the limit we obtain
\begin{equation}
\sum_{i=1}^{N} c^{(i)} \nabla_{\theta^*} \mathcal{L}^{(i)}(\theta^*) = 0.
\end{equation}
This is the first-order necessary condition for Pareto optimality. Therefore, the PAMA algorithm converges to a Pareto stationary point.

\qed

\end{proof}

\newpage
\section{\label{app sec: pseudocode}Pseudocode}
The PAMA algorithm is summarized in \cref{algo:PAMA}. For more details on fine-tuning language models, we strongly recommend that the reader refer to the PPO algorithm~\citep{rlhf} and the open-source TRL package~\citep{vonwerra2022trl}.

\begin{algorithm}
\caption{\underline{PA}reto \underline{M}ulti-Objective \underline{A}lignment (PAMA)}
\label{algo:PAMA}
\begin{algorithmic}[1]
\REQUIRE 
\STATE $\mathcal{D}$: Prompt dataset
\STATE The SFT policy $\pi(\cdot; \theta)$ with parameters $\theta$ 
\STATE Add $N$ new value heads to the language model 
\STATE Set number of iterations $T$ and mini-batch size $B$

\FOR {iteration $t = 1$ to $T$}

    \STATE Sample a mini-batch of prompts $x_j$ from $ \mathcal{D}$.
    
    \STATE For each prompt $x_j$, generate response $y_j \sim \pi(x_j; \theta)$.
    
    \STATE Get $\mathbf{R}(x_j, y_j) = (R^1(x_j, y_j), R^2(x_j, y_j), \dots, R^N(x_j, y_j))$ by reward models.
    
    \STATE Compute the Noon PPO advantage $\hat{A}_j$ (\cref{eq:noon_advantage,app eq: ppo advantage}).
    
    \STATE Solve \cref{eq: our min norm problem} by~\cref{theorem: Optimal Convex Combination for the Scalar Min-Norm Problem}
    
    \STATE Perform gradient ascending using \cref{eq: our min norm problem} to optimize the policy.

    \STATE Optimizing the $N$ value function of Noon PPO~\citep{PPO}.

\ENDFOR

\STATE \Return Optimized policy $\pi$.

\end{algorithmic}
\end{algorithm}

\newpage
\section{~\label{app sec: mgda-ub discussion}Extended Discussion on Related Work}

\subsection{Computational Challenges in Multi-Objective Optimization for LLMs}
Applying MOO to large-scale language models introduces unique computational constraints beyond standard MOO or MORL. Traditional MOO techniques rely on explicit gradient aggregation, which scales poorly when applied to LLMs due to the high dimensionality of their parameter spaces. Gradient-based methods such as MGDA require solving min-norm optimization problems, which incur prohibitive $\mathcal{O}(n^2 d)$ complexity. This makes them impractical when $d$ reaches billions, as is common in modern LLMs.

Recent approaches like ICA propose alternative optimization strategies for multi-task learning. However, their dependence on singular value decomposition introduces numerical instability, particularly when training with low-precision formats (e.g., \texttt{float16}, \texttt{bfloat16}). This issue is further exacerbated under memory constraints when training large-scale models.

\subsection{Multi-Gradient Descent Algorithm and MGDA-UB}

The Multiple Gradient Descent Algorithm is a gradient-based method used in multi-objective optimization to find Pareto-optimal solutions. The goal is to determine a descent direction that minimizes all task losses simultaneously.  

Given $T$ tasks with losses $\hat{L}_t (\theta_{sh}, \theta_t)$, MGDA finds coefficients $\alpha_t$ such that:
\begin{equation}
    \sum_{t=1}^{T} \alpha_t \nabla_{\theta_{sh}} \hat{L}_t (\theta_{sh}, \theta_t) = 0,
\end{equation}
where $\alpha_t \geq 0$ and $\sum_{t=1}^{T} \alpha_t = 1$.  

Since solving this exactly is computationally expensive, the MGDA-UB formulation approximates it by solving:
\begin{equation}
    \min_{\alpha_1, \dots, \alpha_T} \left\| \sum_{t=1}^{T} \alpha_t \nabla_Z \hat{L}_t (\theta_{sh}, \theta_t) \right\|_2^2, \text{ s.t. } \sum_{t=1}^{T} \alpha_t = 1, \quad \alpha_t \geq 0 \quad \forall t.
\end{equation}
Here, $Z = g(X; \theta_{sh})$ represents the shared representation, and $\nabla_Z \hat{L}_t$ denotes the gradients of losses with respect to this representation.  

Theorem 1 in the paper~\citep{mgda_nips} claims that under the assumption that $\frac{\partial Z}{\partial \theta_{sh}}$ is full-rank, the solution $\alpha_1, \dots, \alpha_T$ to MGDA-UB satisfies:
\begin{enumerate}
    \item If $\sum_{t=1}^{T} \alpha_t \nabla_{\theta_{sh}} \hat{L}_t (\theta_{sh}, \theta_t) = 0$, then the current parameters are Pareto stationary.
    \item Otherwise, $\sum_{t=1}^{T} \alpha_t \nabla_{\theta_{sh}} \hat{L}_t (\theta_{sh}, \theta_t)$ is a descent direction that decreases all objectives.
\end{enumerate}

\subsection{MGDA-UB as Baselines}
In our experiments, we use MGDA-UB as a baseline. We outline its derivation below:
\begin{equation}
\label{app eq: upper bound of min norm MGDA-UB}
    \begin{aligned}
          \left\| \sum_{i=1}^{N} c^{(i)} \nabla_{\theta} \mathcal{L}^{(i)} (\theta)  \right\|_2^2 &=  \left\| \sum_{i=1}^{N} c^{(i)} \nabla_{\pi} \mathcal{L}^{(i)} (\theta) \nabla_\theta \pi(\theta) \right\|_2^2 \\
          &=  \left\| \sum_{i=1}^{N} c^{(i)} \frac{1}{\pi_{ref}} I(A^{(i)}) \nabla_\theta \pi(\theta) \right\|_2^2 \\ 
         & \leq  \left\| \sum_{i=1}^{N} c^{(i)}  I(A^{(i)})  \right\|_2^2 \left\| \frac{1}{\pi_{ref}} \nabla_\theta \pi(\theta) \right\|_2^2,
    \end{aligned}
\end{equation}
where 
\begin{equation}
	\label{eq: I A function in MOC}
	I(A) =
	\begin{cases}
		0, & \text{if } (A > 0 \text{ and } u > 1 + \epsilon) \\
		& \quad \text{ or } (A < 0 \text{ and } u < 1 - \epsilon), \\
		A, & \text{if } (A > 0 \text{ and } u \leq 1 + \epsilon) \\
		& \quad \text{ or } (A < 0 \text{ and } u \geq 1 - \epsilon),
	\end{cases}
\end{equation}
\begin{equation}
    \sum_{i=1}^{N} c^{(i)} = 1, \quad c^{(i)} \geq 0  \quad \forall i,
\end{equation}
$A$ is the advantage function of PPO algorithm, and $u=\frac{\pi}{\pi_{ref}}$. 

We first compute the coefficients $c^{(i)}$ by solving:
\begin{equation}
    \min_{c^{(i)}}\left\| \sum_{i=1}^{N} c^{(i)}  I(A^{(i)})  \right\|_2^2.
\end{equation}

These coefficients are then used to minimize the multi-objective loss:
\begin{equation}
    \sum_{i=1}^{N} c^{(i)} \nabla_{\theta} \mathcal{L}^{(i)} (\theta),
\end{equation}
where $\mathcal{L}^{(i)}$ is the PPO loss function. 

\newpage
\section{\label{app sec: Complexity Analysis of Multi Objective Optimization Algorithms}Complexity Analysis of Multi Objective Optimization Algorithms}

In this section, we analyze the computational complexity of the Min-Norm algorithm, a fundamental component of many multi-objective optimization methods~\citep{MGDA,cagrad,pcgrad,ica}. We compare it with our proposed approach to highlight its advantages.

We consider the optimization problem
\begin{equation}
\min_{c \in \Delta} \left\|\sum_{i=1}^{n} c_i x_i\right\|_2^2,
\label{eq:problem}
\end{equation}
where $x_i \in \mathbb{R}^d$ for $i=1,\dots,n$, and the coefficient vector $c = (c_1, c_2, \ldots, c_n)^\top$ is constrained to lie in the probability simplex $ \Delta = \{ c \in \mathbb{R}^n : c_i \ge 0, \sum_{i=1}^{n} c_i = 1 \} $.

Expanding the objective function, we have
\begin{equation}
\left\|\sum_{i=1}^{n} c_i x_i\right\|_2^2 = \left(\sum_{i=1}^{n} c_i x_i\right)^\top \left(\sum_{j=1}^{n} c_j x_j\right) = \sum_{i,j=1}^{n} c_i c_j \, x_i^\top x_j.
\end{equation}

Define the Gram matrix $G \in \mathbb{R}^{n \times n}$ with entries $G_{ij} = x_i^\top x_j$. Then the problem in \eqref{eq:problem} can be reformulated as the following quadratic programming (QP) problem:
\begin{equation}
\min_{c \in \Delta} \; c^\top G c.
\label{eq:qp}
\end{equation}

\subsection{Time Complexity Analysis}

The computational cost of solving \eqref{eq:qp} consists of two main components: (1) constructing the Gram matrix $G$ and (2) solving the QP using an interior-point method.

\subsubsection{Constructing the Gram Matrix}

Each entry $G_{ij}$ is computed as an inner product of two $d$-dimensional vectors, which requires $\mathcal{O}(d)$ operations. Since there are $\mathcal{O}(n^2)$ entries in $G$, the total time complexity for constructing the Gram matrix is
\begin{equation}
\mathcal{O}(n^2 \cdot d).
\end{equation}

\subsubsection{Solving the QP via an Interior-Point Method}

The QP in \eqref{eq:qp} involves $n$ variables. An interior-point method typically requires solving an $n$-dimensional linear system at each iteration, with a computational cost of $\mathcal{O}(n^3)$ per iteration. If the number of iterations needed for convergence is denoted by $I$ (in practice, $I = \mathcal{O}\bigl(\sqrt{n}\,\log(1/\varepsilon)\bigr)$, where $\varepsilon$ is the desired accuracy), then the total cost for solving the QP is
\[
\mathcal{O}(I \cdot n^3) = \mathcal{O}\bigl(n^{7/2}\log(1/\varepsilon)\bigr).
\]

\subsubsection{Overall Time Complexity}

The overall time complexity is the sum of the time to construct $G$ and the time to solve the QP:
\begin{equation}
\mathcal{O}\bigl(n^2 \cdot d + n^{7/2}\log(1/\varepsilon)\bigr).
\end{equation}
Since we are considering scenarios where $d$ is extremely large (e.g., on the order of billions) and $n$ is relatively small, the term $\mathcal{O}(n^2 \cdot d)$ generally dominates the computational cost.

\subsection{Space Complexity Analysis}

The space requirements are determined by the following components:
\begin{itemize}
    \item \textbf{Input Data:} Storing $n$ $d$-dimensional vectors requires $\mathcal{O}(n \cdot d)$ space.
    \item \textbf{Gram Matrix:} The Gram matrix $G$ has $n^2$ entries and requires $\mathcal{O}(n^2)$ space.
    \item \textbf{Additional Storage:} The interior-point method uses extra storage for intermediate variables and factorization matrices, typically contributing at most $\mathcal{O}(n^2)$ additional space.
\end{itemize}
Thus, the total space complexity is
\begin{equation}
\mathcal{O}(n \cdot d + n^2).
\end{equation}
In cases where $d \gg n$, the dominant storage cost is $\mathcal{O}(n \cdot d)$ due to the input data.

\subsection{Discussion}
The complexity analysis highlights that conventional multi-objective optimization algorithms become intractable in large-scale settings, such as LLM training, where $d$ can reach billions. The cost of constructing the Gram matrix and storing high-dimensional data presents a significant computational bottleneck.

In contrast, PAMA achieves both time and space complexity of $\mathcal{O}(n)$, making it highly scalable and efficient.

\newpage
\section{\label{app sec: Additional Details Regarding Experiments}Additional Details Regarding Experiments}
\begin{table}[!hbp]
\centering
\caption{\label{app tab Key configuration of the language models}Key configurations for the experimental setup.}
\begin{tabular}{ll}
\toprule
\textbf{Hyper-parameter} & \textbf{Value} \\
\midrule
\textbf{Computational resources.} & \\             
GPU                                         & A NVIDIA RTX A6000 (48G)                           \\
CPU & Intel(R) Core(TM) i9-14900K \\
Memory & 128 G \\
\midrule 
\textbf{GPT2 (125M) experiments} & \\
Dataset & IMDb~\citep{imbda-dataset} \\
Description & Generate positive sentiment and long review \\
Sentiment model & https://huggingface.co/lvwerra/distilbert-imdb \\
Length reward function &  numpy.clip($l$/140, min=70/140, max=210/140)\\
$l$ & The length of the response string \\
Batch size & 128 \\
Epoch & 1 \\
When to evaluate & After training \\
\midrule 
\textbf{GPT2-XL (1.5B) experiments} & \\
Dataset & Helpful Assistant~\citep{hh-rlhf} \\
Task description                                      & Provide humor and long responses to questions \\
Prompt                                           & Default prompts in the dataset                 \\
Humor reward                                     & \href{https://huggingface.co/mohameddhiab/humor-no-humor}{Humor no humor}                                   \\
Length reward function & numpy.clip($l$/140, 0, 4096/140) \\
$l$ & The length of the response string \\
Evaluation frequency & 50 training steps \\
Epoch & 1 \\

\bottomrule
\end{tabular}
\end{table}
This section provides a detailed description of the experimental setup used to generate the results presented in the main text. Our goal is to ensure transparency, facilitate reproducibility, and support future research based on these findings.

\textbf{Random seeds.} To ensure consistency and reproducibility, we fixed all random seeds across all experiments, including those used in PyTorch, NumPy, Gym, Random, and CUDA. Our implementation is based on~\citet{vonwerra2022trl}.
For experiments in \cref{fig: gpt2_length_quality_comparison}, evaluations were conducted using eight fixed random seeds. For \cref{fig: gpt2xl_humor_length_comparison,fig: llama2_harmless_length_comparison}, a single fixed seed was used due to the computational cost of training LLMs.

\textbf{\cref{fig: gpt2_length_quality_comparison}.} We directly use GPT2 (125M) and GPT2-XL from this website: https://huggingface.co/openai-community/gpt2 without any further fine-tuning. We logged the training data for visualization in~\cref{fig: gpt2_length_quality_comparison,fig: llama2_harmless_length_comparison}. As for~\cref{fig: gpt2xl_humor_length_comparison}, we evaluate the models on the test set every 50 training steps.

\textbf{\cref{fig: llama2_harmless_length_comparison}.} The LLaMA-2 model was initially fine-tuned using Supervised Fine-Tuning (SFT) on positive responses. We then introduced $N$ value heads to the model for further optimization.

\cref{app tab Key configuration of the language models,app tab extended Key configuration of the language models} summarize the key configurations for the experiments.

\begin{table}[!htbp]
\centering
\caption{\label{app tab extended Key configuration of the language models}Extended configurations for experimental settings.}
\begin{tabular}{ll}
\toprule
\textbf{Hyper-parameter} & \textbf{Value} \\
\midrule
\textbf{LLaMA2 (7B) experiments} & \\
Dataset & Helpful Assistant~\citep{hh-rlhf} \\
Task description                                      & Provide harmless and long responses to questions \\
Prompt                                           & Default prompts in the dataset                 \\
Harmless reward                                  & \href{https://huggingface.co/Ray2333/gpt2-large-harmless-reward_model}{gpt2 large harmless reward model}                   \\
Length reward function & numpy.clip($l$/140, 0, 4096/140) \\
$l$ & The length of the response string \\
Epoch & 1 \\
When to evaluate & After training \\
Quantization for training                        & 8bit                                              \\
Fine-tuning                             & LoRA                           \\
LoRA r                                           & 64                                                \\
LoRA alpha                                       & 128                                               \\
LoRA dropout                                     & 0.05                                              \\
Optimizer                                        & Adam                                              \\
Batch size                                       & 32                                                 \\
\midrule
\textbf{SFT in LLaMA2 experiments}                                     & Shared for all algorithms \\
Finetuning steps                                 & 20000                                             \\
Initial learning rate                            & 1.41e-4                                           \\
Learning rate scheduler                          & Linear                                            \\
Fine-tuning                             & LoRA                           \\
LoRA r                                           & 64                                                \\
LoRA alpha                                       & 128                                               \\
LoRA dropout                                     & 0.05                                              \\
Optimizer                                        & Adam                                              \\
Batch size                                       & 64                                                 \\
\midrule
\textbf{RLHF setting} & Shared for all algorithms  \\
KL regularization                                & 0.2                                               \\
Epochs                                           & 1                                                 \\
New value head & $N$ two-layer feed-forward head \\
Units of value head & decoder hidden size \\
Activation of value head & ReLU \\
Learning rate                                    & 1e-5                                              \\
Lambda for GAE                                   & 0.95                                              \\
Gamma                                            & 1                                                 \\
Cliprange                                        & 0.2                                               \\
Number of optimization epochs per batch          & 4                                                 \\
Target KL                                        & 3                                                 \\
\bottomrule
\end{tabular}
\end{table}

\newpage
\section{\label{app sec: Additional Experimental Results}Additional Experimental Results}

In this section, we present additional experimental results to further validate the effectiveness of PAMA in multi-objective alignment.

\cref{app tab: sentiment length reward comparison} provides a quantitative comparison of GPT-2 (125M) using humor and length rewards, showing the mean reward values on the test set. PAMA consistently achieves the highest scores in both sentiment and length, demonstrating its superior performance in balancing multiple objectives.

\begin{figure}[!htbp]
    \centering
    \includegraphics[width=0.5\textwidth]
    {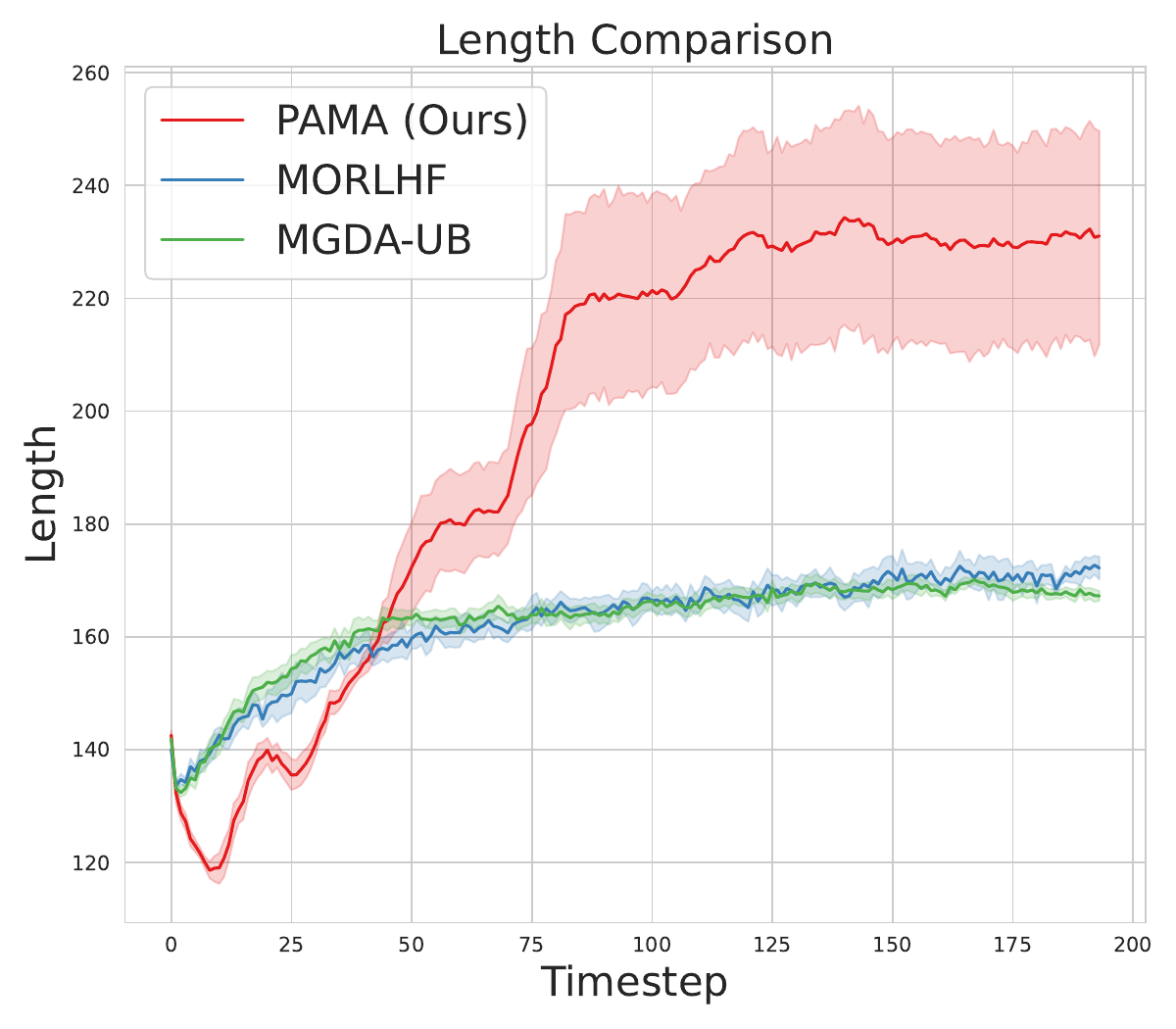}
    \caption{Generation length during training with GPT-2 (125M). Comparison of actual generated text length over the course of training.}
    \label{app fig: gpt2_length}
\end{figure}

\begin{figure}[!htbp]
    \centering
    \includegraphics[width=0.5\textwidth]{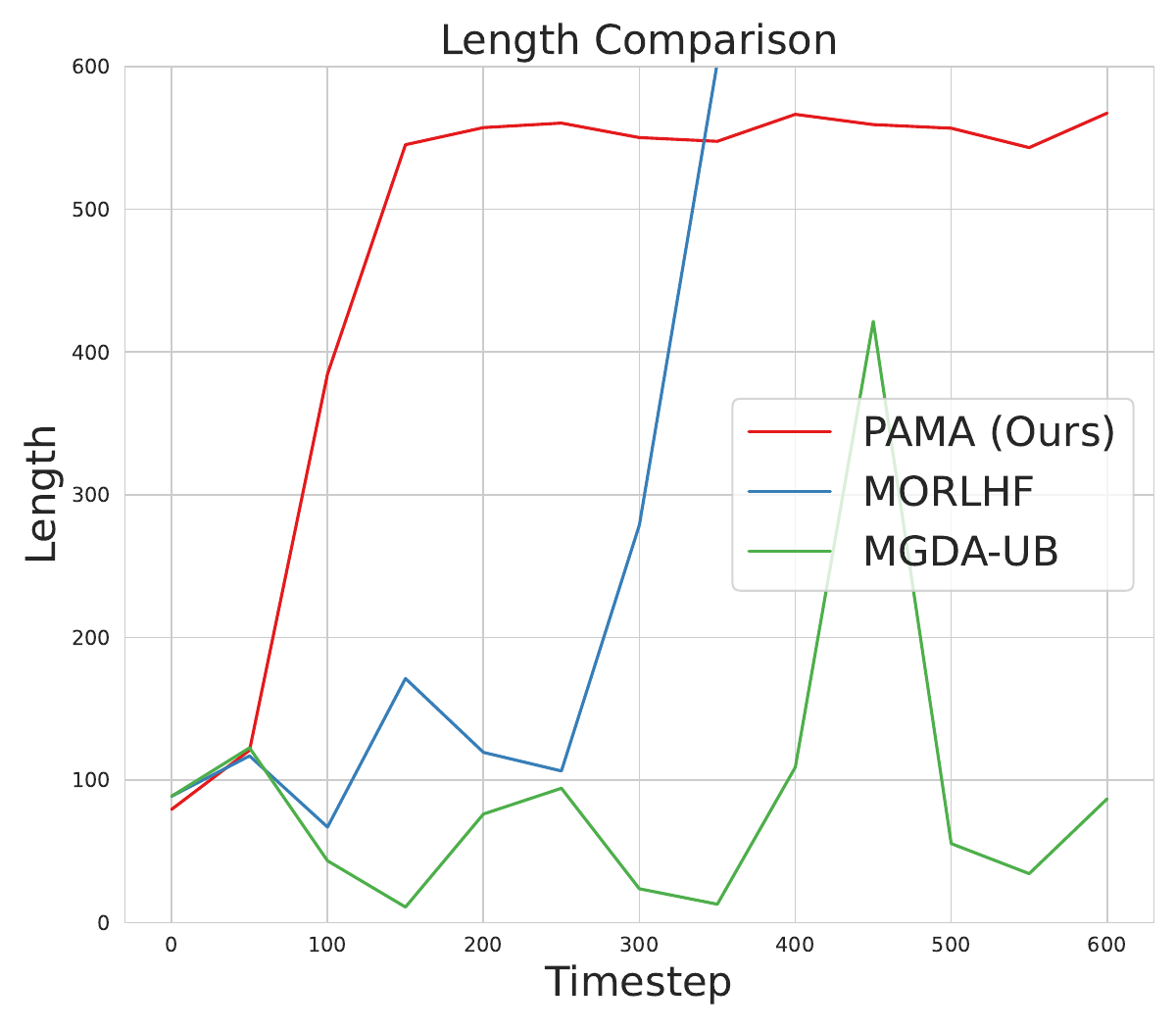}
    \caption{Generation length on the test set with GPT-2 XL (1.5B). The generated text length when using GPT-2 XL (1.5B) as the backbone model.}
    \label{app fig: gpt2_xl_length}
\end{figure}

\cref{app fig: gpt2_length} illustrates the actual generation length of the model during training when using GPT-2 as the backbone. Similarly, \cref{app fig: gpt2_xl_length} presents the generation length on the test set when using GPT-2 XL (1.5B).

\begin{figure*}[!htbp]
    \centering
    \begin{subfigure}{0.48\linewidth}
        \centering
        \includegraphics[width=\linewidth]{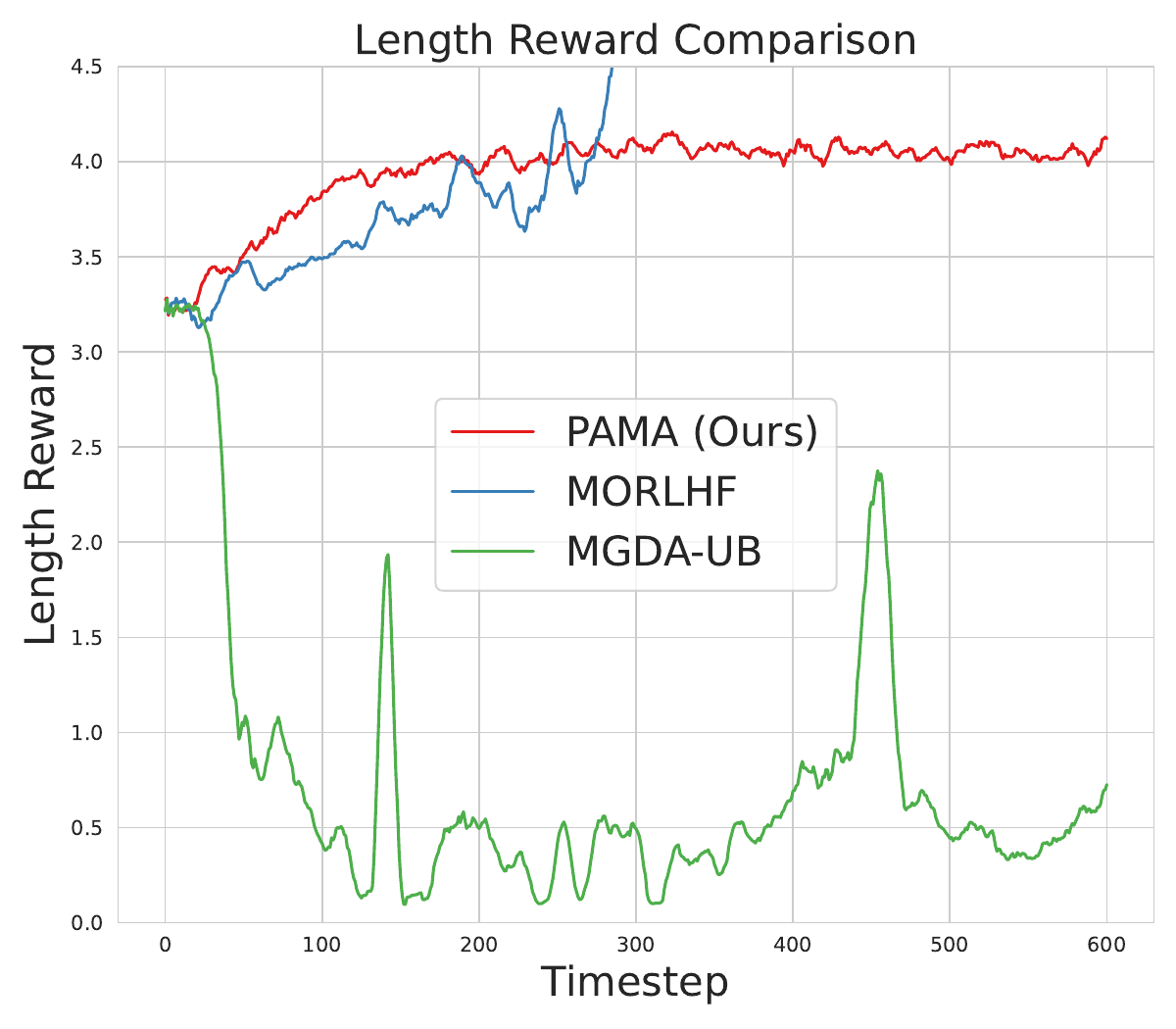}
        \caption{Length reward progression during training}
        \label{app fig: gpt2_length_reward_training}
    \end{subfigure}
    \hfill
    \begin{subfigure}{0.48\linewidth}
        \centering
        \includegraphics[width=\linewidth]{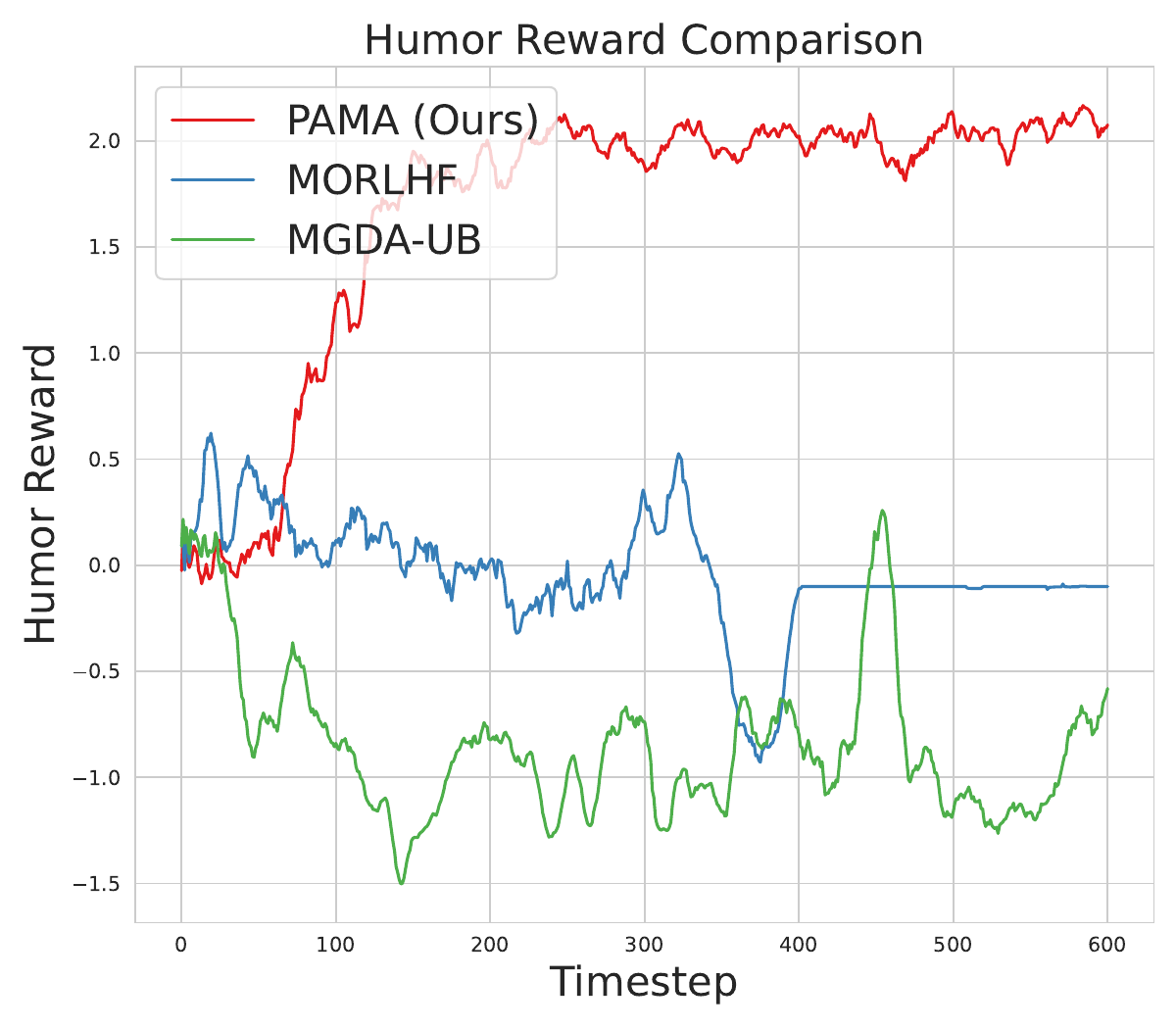}
        \caption{Humor reward progression during training}
        \label{app fig: gpt2_quality_reward_training}
    \end{subfigure}
    \caption{Comparison of length and humor rewards during training with GPT-2 XL (1.5B). PAMA consistently outperforms baseline methods in optimizing both objectives on the HH-RLHF dataset.}
    \label{app fig: gpt2_length_quality_comparison_training}
\end{figure*}

\begin{figure}[!htbp]
    \centering
    \includegraphics[width=0.5\textwidth]{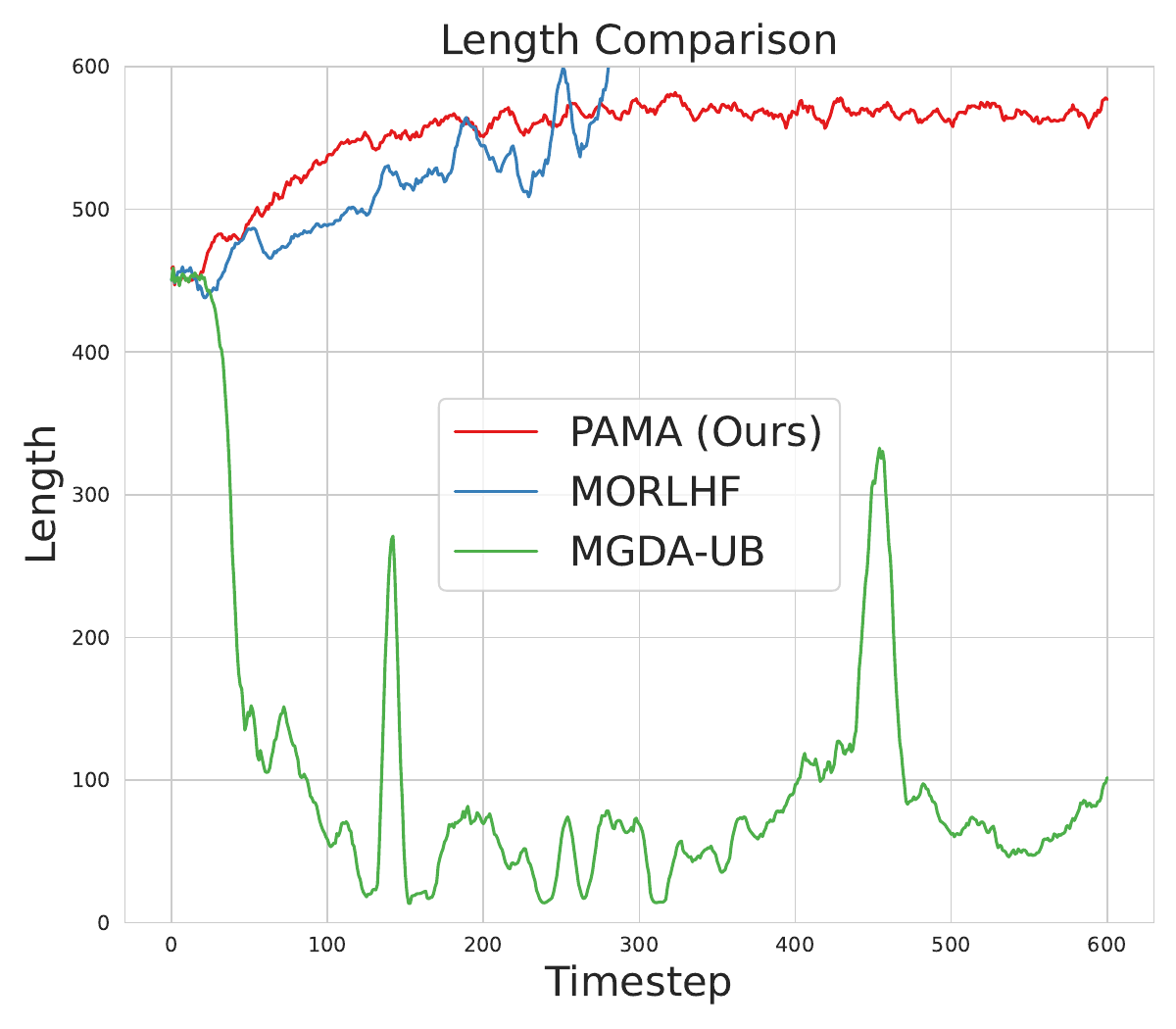}
    \caption{Generation length during training with GPT-2 XL (1.5B). Analysis of actual generated text length over the training process.}
    \label{app fig: gpt2_xl_length_training}
\end{figure}

We further analyze length and quality rewards during training in \cref{app fig: gpt2_length_quality_comparison_training}, where PAMA outperforms baseline methods in balancing these objectives. Additionally, \cref{app fig: gpt2_xl_length_training} shows the actual generation length during training for GPT-2 XL (1.5B).

Finally, \cref{app tab: harmless length reward comparison} presents the harmlessness and length rewards on the test set when using LLaMA2-7B as the base model. The results indicate that MGDA-UB and MORLHF struggle to balance conflicting objectives, whereas PAMA successfully achieves a favorable trade-off.

\begin{table}[!htbp]
    \centering
    \caption{\label{app tab: sentiment length reward comparison}Mean rewards on the test set with GPT-2 (125M). A comparison of sentiment and length rewards for MGDA-UB, MORLHF, and PAMA. PAMA achieves the highest scores across both objectives, demonstrating superior multi-objective alignment.}
    \begin{tabularx}{\textwidth}{lXXX}
        \toprule
        Reward & MGDA-UB & MORLHF & PAMA (Ours) \\
        \midrule
        Sentiment & 1.4734  & 1.6224  & {2.3159}  \\
        \midrule
        Length & 1.1943 & 1.2288 & {1.4179}  \\
        \bottomrule
    \end{tabularx}
    \vspace{-0.3in}
\end{table}

\begin{table}[h]
    \centering
    \caption{\label{app tab: harmless length reward comparison}Mean rewards (harmlessness and length) on the test set with LLaMA2-7B. MGDA-UB and MORLHF struggle to balance conflicting objectives, whereas PAMA effectively optimizes both harmlessness and length.}
    \begin{tabularx}{\textwidth}{lXXX}
        \toprule
        Reward & MGDA-UB & MORLHF & PAMA (Ours) \\
        \midrule
        Harmlessness & -0.2844  & -0.1313  & {0.4406}  \\
        \midrule
        Length & {9.0842}  & 5.3502 & 3.1400 \\
        \bottomrule
    \end{tabularx}
\end{table}


\end{document}